\documentclass[10pt,twocolumn,letterpaper]{article}
\usepackage[pagenumbers]{cvpr} 

\newcommand{\algohighlight}[1]{\textcolor{blue}{#1}}

\usepackage{graphicx}
\usepackage{subcaption}
\usepackage{tikz}
\usepackage[table]{xcolor}
\usepackage{multirow}
\usepackage{makecell}

\usepackage{amsmath}
\usepackage{booktabs}
\usepackage{bm}
\usepackage[ruled,linesnumbered]{algorithm2e}
\usepackage{algpseudocode}
\usepackage{color}
\usepackage{xcolor}

\newcommand{\method}{{\color[RGB]{0,0,0}\texttt{VDOT}}\xspace}
\newcommand{\methodbench}{{\color[RGB]{0,0,0}\texttt{UVCBench}}\xspace}
\newcommand{\methodbf}{{\color[RGB]{0,0,0}\textbf{\texttt{VDOT}}}\xspace}

\definecolor{tabhighlight}{HTML}{e5e5e5}
\definecolor{video_quality}{HTML}{F0D695}
\definecolor{user_study}{HTML}{82ACD1}
\definecolor{score_average}{HTML}{98B567}

\usepackage{utfsym}

\definecolor{cvprblue}{rgb}{0.21,0.49,0.74}
\usepackage[pagebackref,breaklinks,colorlinks,allcolors=cvprblue]{hyperref}

\title{VDOT: Efficient Unified Video Creation via Optimal Transport Distillation}

\author{
Yutong Wang$^{1}$\quad Haiyu Zhang$^{3,2}$\quad Tianfan Xue$^{4,2}$\quad Yu Qiao$^{2}$\\ Yaohui Wang$^{2}$\footnotemark[1] \quad Chang Xu$^{1}$\thanks{Corresponding author.}  \quad Xinyuan Chen$^{2}$\footnotemark[1]\\
$^1$USYD \quad $^2$Shanghai AI Laboratory \quad $^3$BUAA  \quad $^4$CUHK \\
{\tt Project Page: \url{https://vdot-page.github.io/}}
}

\begin{document}
\maketitle
\begin{abstract}
The rapid development of generative models has significantly advanced image and video applications. 
Among these, video creation, aimed at generating videos under various conditions, has gained substantial attention. 
However, existing video creation models either focus solely on a few specific conditions or suffer from excessively long generation times due to complex model inference, making them impractical for real-world applications. 
To mitigate these issues, we propose an efficient unified video creation model, named \method. 
Concretely, we model the training process with the distribution matching distillation (DMD) paradigm. 
Instead of using the Kullback-Leibler (KL) minimization, we additionally employ a novel computational optimal transport (OT) technique to optimize the discrepancy between the real and fake score distributions. 
The OT distance inherently imposes geometric constraints, mitigating potential zero-forcing or gradient collapse issues that may arise during KL-based distillation within the few-step generation scenario, and thus, enhances the efficiency and stability of the distillation process. 
Further, we integrate a discriminator to enable the model to perceive real video data, thereby enhancing the quality of generated videos. 
To support training unified video creation models, we propose a fully automated pipeline for video data annotation and filtering that accommodates multiple video creation tasks. 
Meanwhile, we curate a unified testing benchmark, \methodbench, to standardize evaluation.
Experiments demonstrate that our 4-step \method outperforms or matches other baselines with 100 denoising steps. 
\end{abstract}

\section{Introduction}
\label{sec:intro}

Recent years have witnessed the remarkable development of AI-generated content (AIGC).
Particularly in image and video generation, the impressive capability to generate high-quality perceptual data has attracted interest from both academia and industry. 
The ability to synthesize and edit media under diverse conditions has significantly broadened the scope of application for generative models, facilitating their integration across a wide range of domains.

Despite progress, most existing generative models remain task-specific, with models tailored to narrowly defined objectives. 
While unified image generation and editing have seen notable breakthroughs~\cite{xiao2025omnigen,tan2025ominicontrol,xia2025dreamomni,wang2024genartist,qin2025lumina}, attempts at unified video creation remain comparatively limited~\cite{jiang2025vace,ye2025unic,mou2025instructx}. 
Recently, VACE~\cite{jiang2025vace} reformulates various conditioning signals into unified frame and mask representations and introduces additional adapters for context processing. 
Similarly, UNIC~\cite{ye2025unic} proposed a unified token-based framework that encodes all inputs into three token categories, combined with native attention and task-aware rotary position embeddings (RoPE) to differentiate tasks. 
Although these methods achieve impressive visual fidelity, their architectural complexity and large parameter budgets result in substantial inference latency, limiting their practicality in real-world deployments.

To address the above challenges, we propose a novel distillation framework for unified video creation based on computational optimal transport (OT) techniques. 
Specifically, we formulate the distillation within the distribution-matching distillation paradigm. 
Instead of solely relying on the conventional reverse KL divergence between the teacher and student score distributions, we incorporate an OT-based discrepancy that enforces geometrically meaningful alignment. 
This constraint regularizes the transport direction and effectively mitigates the collapse problem that often arises in few-step distillation. 
Furthermore, we employ an adversarial discriminator that leverages real videos to improve fidelity and counteract undesirable biases inherited from the foundation-scale video models. 
We adopt an alternating optimization scheme to jointly update the generator and critics, yielding a few-step unified video creator with improved visual quality and efficiency.

Due to the scarcity of large-scale open-source datasets and evaluations for unified video creation, we additionally construct a comprehensive multi-task training dataset and evaluation benchmark, termed \methodbench. 
The construction pipeline is fully automated, including 4K-resolution video collection, dense captioning via vision-language models, task-aware data filtering, and candidate ranking, ensuring high-quality and diverse samples that support unified conditioning.
To enable standardized and scalable evaluation, \methodbench supports 18 generation tasks, each with 20 representative test cases spanning a broad range of video types. 
We hope that this automated data construction pipeline and unified benchmark will advance future research in this area.

In summary, the contributions of this work are two-fold: 
\begin{itemize}
    \item We propose \method, an efficient unified video creation framework based on optimal-transport distillation. 
    The OT regularizer provides a geometric constraint to distribution matching, improving training stability and efficiency. 
    To the best of our knowledge, this is the first application of OT within distribution-matching distillation.
    \item We develop a fully automated multi-task data construction pipeline and curate a comprehensive benchmark, \methodbench. 
    Experiments on \methodbench demonstrate that our unified video creator achieves superior performance on both objective metrics and human evaluations while maintaining few-step inference.
\end{itemize}

\section{Related Works}
\label{sec:relatedworks}

\subsection{Visual Creation and Editing}
The rapid advancement of image~\cite{chen2023pixart,esser2024scaling,saharia2022photorealistic,li2024hunyuan} and video~\cite{ma2024latte,wang2025lavie,kong2024hunyuanvideo,wan2025wan,hacohen2024ltx} generation models has significantly impacted advertising, film production, e-commerce, and interactive entertainment~\cite{pan2023drag,wang2024instantid}. 
To meet diverse and personalized application needs, numerous methods for precise control and editing have emerged~\cite{tan2025ominicontrol,zhang2023magicbrush}. 
Most of these approaches produce high-quality visuals conditioned on pose, depth, optical flow, or reference images. 
Conditional image generation is commonly enabled via ControlNet~\cite{zhang2023adding} or T2I-Adapter~\cite{mou2024t2i}, while OmniGen~\cite{xiao2025omnigen} and OmniControl~\cite{tan2025ominicontrol} extend to multi-task image editing and generation. 
By contrast, most video editing systems remain specialized, such as animating characters~\cite{dai2023animateanything,hu2024animate}, canvas outpainting~\cite{dehan2022complete}, and colorization~\cite{zhang2019deep}. 
Recent efforts toward all-in-one video creation and editing, such as VACE~\cite{jiang2025vace} and UNIC~\cite{ye2025unic}, show promise but suffer from complex architectures and large parameter counts, leading to long processing times. 
In this paper, we build on VACE and distill it into a few-step generator; moreover, by integrating a discriminator, we suppress undesirable artifacts and biases present in the base model.

\subsection{Visual Distillation}
In visual distillation, several influential approaches have recently emerged, including Progressive Distillation~\cite{salimans2022progressive}, Consistency Distillation~\cite{geng2024consistency,kim2023consistency,wang2024phased}, Score Distillation~\cite{katzir2023noise,wei2024adversarial}, Rectified Flow~\cite{liu2022flow,liu2023instaflow,yan2024perflow}, and Adversarial Distillation~\cite{lu2025adversarial,ge2025senseflow,sauer2024adversarial}. 
These methods typically train a student generator to follow the teacher’s ODE-defined sampling trajectory using substantially fewer steps. 
Distribution matching distillation (DMD)~\cite{yin2024one,yin2024improved} acts in another way. 
It extends the score distillation by minimizing the expectation over $t$ of approximate KL divergences between the teacher’s diffused data distribution and the student’s diffused output distribution. 
Self-Forcing~\cite{huang2025self} extends the DMD paradigm to the video by using a DMD objective and a denoising loss to distill a few-step video generator. 
However, in the few-step regime, relying solely on DMD loss can lead to zero-forcing and gradient collapse, resulting in unstable training and susceptibility to model-seeking problems. 
To address these issues, we propose a novel distribution matching objective that imposes geometric constraints on distribution alignment, enhancing training stability and accelerating convergence.

\subsection{Optimal Transport in Computer Vision}
The computational optimal transport (OT) techniques~\cite{peyre2019computational} have become increasingly popular tools in computer vision for aligning probability distributions. 
These methods have been successfully applied to point cloud registration~\cite{bonneel2019spot}, image generation~\cite{tartavel2016wasserstein, dukler2019wasserstein}, as well as video understanding~\cite{luo2022weakly, wang2023self, wang2024inverse} and generation~\cite{acharya2018towards}. 
Many challenging optimization problems can be cast as minimizing the OT distance. 
In high-dimensional generative modeling, minimizing the Wasserstein distance between data and model distributions motivates the Wasserstein autoencoders~\cite{tolstikhin2017wasserstein}, while maximizing the Kantorovich dual yields the Wasserstein GAN (WGAN)~\cite{arjovsky2017wasserstein}. 
Beyond generative modeling, the detector YOLOX adopts Optimal Transport Alignment (OTA) for label assignment, exploiting OT’s strong matching behavior~\cite{ge2021yolox}.
These successes motivate an OT-based objective for distribution matching for few-step video distillation.

\section{Preliminary}
\subsection{Multimodal Inputs and Video Condition Unit} 

Existing video editing and creation tasks differ in objectives and input forms, yet all can be represented through four basic modalities: \textbf{text}, \textbf{image}, \textbf{video}, and \textbf{mask}. 
All creation tasks can be categorized into five classes: \textbf{Text-to-Video Generation (T2V)}, \textbf{Reference-to-Video Generation (R2V)}, \textbf{Video-to-Video Generation (V2V)}, \textbf{Masked Video-to-Video Generation (MV2V)}, and \textbf{composite tasks}. 
Following VACE~\cite{jiang2025vace}, we introduce the Video Condition Unit (VCU): 
\begin{equation}\label{eqn:vcu}
   V=[T;F;M],
\end{equation}
where $T$ is a text prompt, $F=\{u_1,\ldots,u_n\}$ denotes context frames, and $M=\{m_1,\ldots,m_n\}$ consists of aligned binary masks. 
Each frame $u_i \in [-1,1]^{3\times h\times w}$ and mask $m_i \in \{0,1\}^{h\times w}$ indicate editable regions, in which ``1"s and ``0"s symbolize where to edit or not. 
As illustrated in Table~\ref{tab:vcu}, by adjusting the frames and masks, the VCU generalizes to all video tasks.

\subsection{Wasserstein Discrepancy}
Given two distributions \(\bm{A} \in \mathbb{R}^{I \times D}\) and \(\bm{B} \in \mathbb{R}^{J \times D}\), an effective way to measure their difference is through the sample-based Wasserstein discrepancy~\cite{cuturi2013sinkhorn}, defined as:
\begin{eqnarray}\label{eqn:wd}
\begin{aligned}
 \mathbb{W}_{2}(\bm{A}, \bm{B})
&= \min_{\bm{T} \in \Pi(\bm{u}, \bm{\mu})} \mathbb{E}_{(\bm{a}, \bm{b}) \sim \bm{T}}[d(\bm{a}, \bm{b})] \\
&= \min_{\bm{T} \in \Pi(\bm{u}, \bm{\mu})} \langle \bm{D}_{ab},~ \bm{T} \rangle,
\end{aligned}
\end{eqnarray}
where \(\bm{D}_{ab} = [d(\bm{a}_i, \bm{b}_j)] \in \mathbb{R}^{I \times J}\) is a distance matrix, with each element \(d(\bm{a}_i, \bm{b}_j)\) representing the distance between the \(i\)-th sample from \(\bm{A}\) and the \(j\)-th sample from \(\bm{B}\). For the Wasserstein distance, we typically apply the Euclidean distance matrix. 
The set \(\Pi(\bm{u}, \bm{\mu}) = \{ \bm{T} \geq 0 \mid \bm{T} \bm{1}_J = \bm{u}, \bm{T}^T \bm{1}_I = \bm{\mu} \}\) represents the set of doubly-stochastic matrices, where the marginals must lie on the Simplex, i.e., \(\bm{u} \in \Delta^{I-1}\) and \(\bm{\mu} \in \Delta^{J-1}\). Generally, we set the marginals to be uniform, i.e., \(\bm{u} = \frac{1}{I} \bm{1}_I\) and \(\bm{\mu} = \frac{1}{J} \bm{1}_J\).
The optimal transport matrix corresponding to \(\mathbb{W}_{2}(\bm{A}, \bm{B})\), denoted as \(\bm{T}^* = [\bm{t}_{ij}^*]\), is the optimal joint distribution of the samples and targets that minimizes the expectation of the distance.

\begin{table}[t]
\caption{
The representation of frames ($F$s) and masks ($M$s) under the four basic tasks~\cite{jiang2025vace}.
}
\setlength\tabcolsep{15pt}
\begin{tabular}{c|c}
\toprule
Tasks  &  Frames ($F$s) \& Masks ($M$s) \\ \midrule 
\multirow{2}{*}{T2V}  
                      & $F=\{0_{h \times w}\} \times n$ 
\\
                      & $M=    \{1_{h \times w}\} \times n$ 
\\ \midrule 
\multirow{2}{*}{R2V}  
                      & $F=    \{r_1, r_2,...,r_l\} + \{0_{h \times w}\} \times n$ 
\\
                      & $M=    \{0_{h \times w}\} \times l + \{1_{h \times w}\} \times n$ 
\\ \midrule 
\multirow{2}{*}{V2V}  
                      & $F=\{u_1,u_2,...,u_n\}$                                                                  
\\
                      & $M=    \{1_{h \times w}\} \times n$
\\ \midrule 
\multirow{2}{*}{MV2V} 
                      & $F=\{u_1,u_2,...,u_n\}$                                                                  
\\
                      & $M=\{m_1,m_2,...,m_n\}$                                                                 
\\ 
\bottomrule
\end{tabular}
\label{tab:vcu}
\end{table}

\subsection{Distribution Matching Distillation}
Distribution Matching Distillation (DMD)~\cite{yin2024one,yin2024improved} is a method for distilling pretrained diffusion models $\bm{F}_\psi$ into efficient one-step or multi-step generators $\bm{G}_\theta$ by minimizing the reverse Kullback-Leibler (KL) divergence between the teacher (real) distribution $\bm{p}_{\text{real}}$ and the student (fake) distribution $\bm{p}_{\text{fake}}$ generated by the model. The reverse KL divergence is given by:
\begin{equation}\label{eqn:kl}
   \mathbb{D}_{\text{KL}}(\bm{p}_{\text{fake}} \| \bm{p}_{\text{real}}) = \int \bm{p}_{\text{fake}}(\bm{x}) \log \frac{\bm{p}_{\text{fake}}(\bm{x})}{\bm{p}_{\text{real}}(\bm{x})} \, d\bm{x},
\end{equation}
This divergence quantifies the information lost when $\bm{p}_{\text{fake}}$ is used to approximate $\bm{p}_{\text{real}}$, and minimizing it aligns the generated distribution with the real data distribution. 
The gradient of the DMD objective with respect to the generator parameters $\theta$ is given by: 
\begin{equation}
   \nabla_\theta \mathcal{L}_{\text{DMD}} = \mathbb{E}_{\bm{z}, t', t, \bm{x}_t} \left[ (\bm{s}_{\text{real}}(\bm{x}_t) - \bm{s}_{\text{fake}}(\bm{x}_t)) \frac{d\bm{G}_\theta(\bm{z}, t')}{d\theta} \right],
\end{equation}
where $\bm{z} \sim \mathcal{N}(\bm{0}, \bm{I})$ is a random latent variable. 
$t' \sim \mathcal{U}(0, T)$ is randomly selected from the generator schedule, and 
$\bm{x}_t$ is the noisy sample obtained by diffusing the generator output $\hat{\bm{x}}_0$, $\bm{x}_t = \bm{q}(\bm{x}_t | \hat{\bm{x}}_0)$. 
The real score $\bm{s}_{\text{real}}(\bm{x}_t)$ and the fake score $\bm{s}_{\text{fake}}(\bm{x}_t)$ are the gradients of the log probabilities of the real and fake distributions, respectively:
\begin{equation}
   \bm{s}_{\text{\{real / fake\}}}(\bm{x}_t) = \nabla_{\bm{x}_t} \log \bm{p}_{\text{\{real / fake\}}}(\bm{x}_t).
\end{equation}

The final DMD loss is computed as:
\begin{equation}\label{eqn:dmd}
   \mathcal{L}_{\text{DMD}}(\theta) = \mathbb{E}_{\bm{z}, t, \bm{x}_t} \left[ \|\hat{\bm{x}}_0 - \text{sg}(\hat{\bm{x}}_0 - \nabla_{\text{KL}}(\bm{x}_t, t))\|_2^2 \right],
\end{equation}
where $\text{sg}(\cdot)$ denotes the stop-gradient operation. 
In practice, the gradient $\nabla_{\text{KL}}$ can be approximated by the minus of score functions, $\nabla_x \mathbb{D}_{\text{KL}}(\bm{p}_{\text{fake}} \| \bm{p}_{\text{real}}) \approx \bm{s}_{\text{fake}}(x) - \bm{s}_{\text{real}}(x)$.

\begin{figure*}[t]
    \centering
    \includegraphics[width=1.02\textwidth]{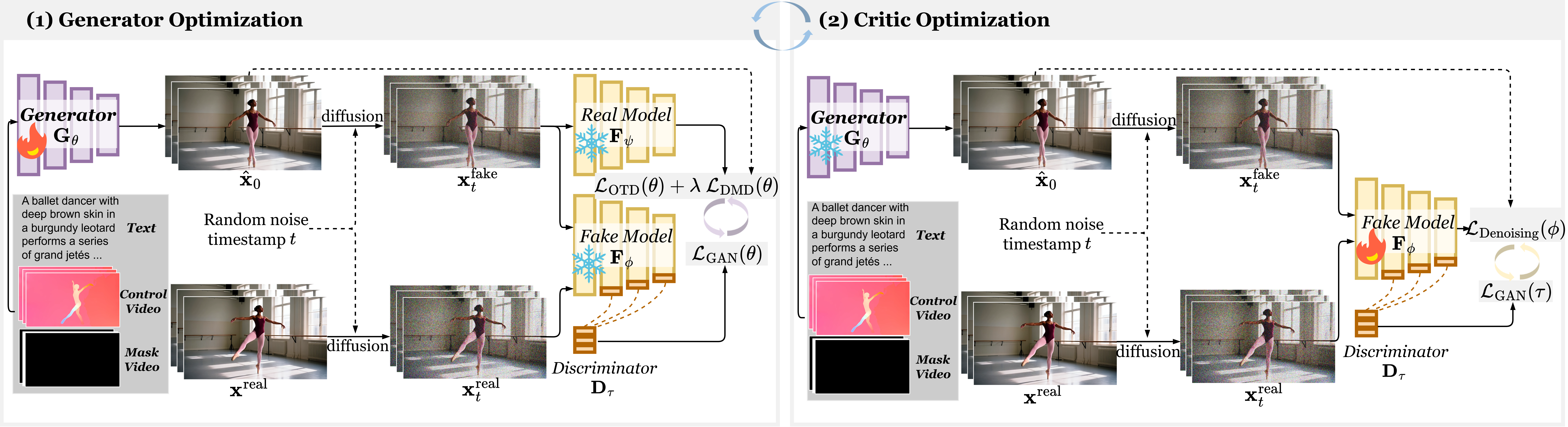}
    \caption{\textbf{The training pipeline of \method.} 
    By integrating the OTD loss, DMD loss, and GAN loss, we distill the teacher model into a few-step unified video creator. 
    At each step, we alternately train the generator and the critics, while between steps, we alternate between distribution matching and adversarial objectives. 
    }
    \label{fig:schema}
    \vspace{-10pt}
\end{figure*}

\section{Methodology}
\label{sec:methods}

\subsection{Overview}

\method is designed as an efficient unified video creator that accepts text, images, videos, and masks as inputs and produces a task-compliant output video with few denoising steps. 
We adopt the pretrained VACE-Wan2.1-14B~\cite{jiang2025vace} as the base generator, a state-of-the-art all-in-one model for video creation and editing. 
VACE comprises frozen Wan DiT blocks and newly introduced VACE DiT blocks; the latter process contextual information derived from conditional inputs. 
A Video Condition Unit (VCU) converts heterogeneous conditioning signals into a unified representation. 
Inputs are then preprocessed and tokenized into video, text, and context tokens. 
Wan blocks process the video tokens, while VACE blocks handle the context tokens; the outputs of the VACE blocks are fused into the corresponding layers of the Wan backbone. 
Further architectural details can be found in the VACE paper~\cite{jiang2025vace}. 
For brevity, we omit the explicit notation for text and context tokens in the following sections.

To enable few-step video generation, we propose a computational optimal-transport (OT)–based step-distillation framework. 
Specifically, we cast training of the few-step generator $\bm{G}_{\theta}$ within the Distribution-Matching Distillation (DMD) paradigm, employing two score networks, $\bm{F}_{\psi}$ and $\bm{F}_{\phi}$, that estimate the teacher (real) and student (fake) score distribution, respectively. 
Instead of relying solely on KL minimization, we incorporate an OT discrepancy to constrain the distribution matching geometrically. 
We further augment the fake model with an adversarial discriminator to correct score-approximation errors and mitigate biases inherited from the base model by leveraging real video data. 
Within each step, we alternately train the generator and the critics, while between steps, we alternate between distribution matching objectives and adversarial objectives. 
An overview of the framework is illustrated in Figure~\ref{fig:schema}.

\subsection{Optimal Transport Distillation}
The objective of DMD is to minimize the reverse KL divergence, $\mathbb{D}_{\text{KL}}(\bm{p}_{\text{fake}} \| \bm{p}_{\text{real}})$. 
The fake score function enables backpropagation, allowing the generator $\bm{G}_{\theta}$ to update and move the student  distribution closer to the teacher distribution. 
However, a common issue that arises is model seeking. 

As shown in Equation~\eqref{eqn:kl}, reverse KL tends to overemphasize regions of the target distribution where the model has high probability, often ignoring areas of the target distribution with lower probability. 
This leads the model to seek and overfit specific regions. 
In the few-step generation scenario, this issue is exacerbated. 
For example, when training a few-step generator, the difference between the real and fake score distributions is initially large. 
Without directional guidance, model training is prone to encountering zero-forcing or gradient collapse, ultimately failing to capture the diversity of the target distribution and generalizing poorly across the entire distribution, as shown in Figure~\ref{fig:ot_illu}.
\begin{itemize}
    \item For any point in $\{x\ |\ \bm{p}_{\text{fake}}(x)\to0 \ \text{and} \ \bm{p}_{\text{real}}(x)>0  \}$, the integrated $\mathbb{D}_{\text{KL}}\to 0$. 
    This causes the models to neglect regions where $\bm{p}_{\text{real}}(x)>0$, preventing those regions from updating. This is the so-called zero-forcing problem~\cite{lu2025adversarial}. 
    As a result, the student distribution fails to fully cover the teacher distribution. 
    \item For any point in $\{x\ |\ \bm{p}_{\text{fake}}(x)>0 \ \text{and} \ \bm{p}_{\text{real}}(x)\to 0  \}$, the integrated $\mathbb{D}_{\text{KL}}\to +\infty$, leading to gradient collapse or training instability.
\end{itemize}

\begin{figure}[t]
    \centering
    \includegraphics[width=0.50\textwidth]{figures/ot_illu.png}
    \caption{Illustration of potential problems caused by reverse KL divergence and the strength of optimal transport constraint. 
    }
    \label{fig:ot_illu}
    \vspace{-14pt}
\end{figure}

The work in ADP~\cite{lu2025adversarial} addresses this issue by pre-training the fake model via adversarial optimization, thereby alleviating the mode-seeking problem. 
However, this paradigm requires collecting numerous ODE pairs from the offline teacher model and generating noisy samples through interpolation, which is both costly and labor-intensive.

In this paper, we introduce a computational optimal transport discrepancy as a geometric constraint to aid the optimization of the generator $\bm{G}_\theta$. 
The OT discrepancy calculates the minimum transport cost between two distributions and the corresponding optimal transport plan, establishing a one-to-one correspondence. 
Concretely, for two score distributions $\bm{p}_{\text{fake}} =[a_i]\in \mathbb{R}^{I \times D} $ and $\bm{p}_{\text{real}}=[b_j]\in \mathbb{R}^{J \times D}$, we apply the entropic optimal transport (EOT) discrepancy: 
\begin{eqnarray}\label{eqn:eot}
\begin{aligned}
 \mathbb{W}_{2}^{\epsilon}(\bm{p}_{\text{fake}}, \bm{p}_{\text{real}})
 = \min_{\bm{T} \in \Pi(\bm{u}, \bm{\mu})} \langle \bm{D},~ \bm{T} \rangle + \epsilon \underbrace{\langle \bm{T}, \log \bm{T} \rangle}_{\text{Entropy Term}}, 
\end{aligned}
\end{eqnarray}
where $\epsilon$ controls the intensity of the entropy term. 
The EOT problem can be solved efficiently using the Sinkhorn algorithm~\cite{cuturi2013sinkhorn} with complexity $\mathcal{O}(IJ)$. 

According to the envelope theorems~\cite{milgrom2002envelope}, the derivative of the objective function with respect to $\bm{D}$ is the optimal transport plan $\bm{T}^*$, $\frac{\partial \mathbb{W}_{2}^{\epsilon}}{\partial \bm{D}}=\bm{T}^*$. 
When the distance matrix is defined by Euclidean distance, $\bm{d}(\bm{a}_i,\bm{b}_j)=\frac{1}{2}\| \bm{a}_i-\bm{b}_j \|^2$, the gradient of the objective with respect to any $\bm{a}_i$ is: 
\begin{equation}
\nabla_{\bm{a}_i}\mathbb{W}_{2}^{\epsilon}=\sum_{j}\bm{T}_{ij}^*\nabla_{\bm{a}_i}\frac{1}{2}\| \bm{a}_i-\bm{b}_j \|^2=\sum_{j}\bm{T}_{ij}^*(\bm{a}_i-\bm{b}_j),
\end{equation}
We can then compute the gradient of the objective with respect to the noisy sample $\bm{x}_t$: 
\begin{equation}\label{eqn:otg}
\nabla_{\text{OT}}(\bm{x}_t, t)=\nabla_{\bm{x}_t}\mathbb{W}_{2}^{\epsilon}=\sum_{ij}\frac{\partial}{\partial \bm{x}_t}\left( \bm{T}_{ij}^*(a_i-b_j) \right).
\end{equation}
In practice, we compute the gradient $\nabla_{\text{OT}}(\bm{x}_t, t)$ through \texttt{torch.autograd}. 
Finally, the optimal transport distillation (OTD) loss is computed in the same manner as the DMD loss: 
\begin{equation}\label{eqn:otd}
   \mathcal{L}_{\text{OTD}}(\theta) = \mathbb{E}_{\bm{z}, t, \bm{x}_t} \left[ \|\hat{\bm{x}}_0 - \text{sg}(\hat{\bm{x}}_0 - \nabla_{\text{OT}}(\bm{x}_t, t))\|_2^2 \right].
\end{equation}

\subsection{Generative Adversarial Networks}
The framework described above does not employ the Teacher-Forcing paradigm~\cite{jinpyramidal,zhang2025test}, which typically requires real video data as denoising conditions. 
Instead, following the Self-Forcing approach~\cite{huang2025self}, it uses previously denoised frames to denoise the current frame, thus maintaining consistency between training and testing. 
However, relying solely on distribution matching objectives, without access to real data, leads to approximation errors in the real score function $\bm{F}_{\psi}$~\cite{yin2024improved}, which manifest as artifacts in video textures and details. 
Moreover, without real data inputs, the output quality of the generator $\bm{G}_{\theta}$ is constrained by the teacher model, while also learning some of the teacher's undesirable prototypes, as discussed further in the Appendix. 

To address this limitation, we introduce real data and a discriminator to correct the score functions by incorporating the generative adversarial networks (GANs) objective. 
Specifically, we select three blocks, blocks 23, 31, and 39, from the denoised blocks of the fake score function $\bm{F}_{\phi}$. 
We introduce three learnable registration tokens that interact with the corresponding blocks through cross-attention. 
The resulting outputs are concatenated along the channel dimension and passed through a linear layer-based classifier that outputs classification logits. 
We denote the involved registration tokens, cross-attention blocks, and classifier as the discriminator $\bm{D}_{\tau}$, with parameter $\tau$. 
Given a real video corresponding to the input prompt, we first encode it into the same latent space using a pre-trained VAE, denoted as $\bm{x}^{\text{real}}$. 
Then, using the randomly sampled timestamp from the scheduler, we add noise to $\bm{x}^{\text{real}}$ and $\bm{\hat{x}}_0$, yielding $\bm{x}_t^{\text{real}}$ and $\bm{x}_t^{\text{fake}}$, respectively. 
A relative GAN loss is used to calibrate the score functions: 
\begin{eqnarray}\label{eqn:gan1}
\small 
\mathcal{L}_{\text{GAN}}(\theta) = \underset{\bm{z},t}{\mathbb{E}} \left[ -\left( \bm{D}_{\tau}\left( \bm{x}_t^{\text{fake}},t \right) - \bm{D}_{\tau}\left( \bm{x}_t^{\text{real}},t \right) \right) \right],
\end{eqnarray}
\begin{eqnarray}\label{eqn:gan2}
\small 
\mathcal{L}_{\text{GAN}}(\tau) = \underset{\bm{x}^{\text{fake}}_t,t}{\mathbb{E}} \left[ -\left( \bm{D}_{\tau}\left( \bm{x}_t^{\text{real}},t \right) - \bm{D}_{\tau}\left( \bm{x}_t^{\text{fake}},t \right) \right) \right].
\end{eqnarray}

\subsection{Model Learning}

Our training employs an alternating strategy to optimize the generator and critics. 
The parameters of the real score $\bm{F}_{\psi}$ remain frozen throughout training. 
At each step, we first freeze the fake model $\bm{F}_{\phi}$ and train the generator $\bm{G}_{\theta}$ using the distribution matching objective $\mathcal{L}_{\text{OTD}}(\theta)+\lambda \mathcal{L}_{\text{DMD}}(\theta)$ or adversarial objective $\mathcal{L}_{\text{GAN}}(\theta)$. 
Then, we freeze the generator and train both the fake model $\bm{F}_{\phi}$ and the discriminator $\bm{D}_{\tau}$ using either the diffusion denoising objective $\mathcal{L}_{\text{Denoising}}(\phi)$ or the adversarial objective $\mathcal{L}_{\text{GAN}}(\tau)$. 

\begin{algorithm}
\caption{Training Algorithm of \method.}
\label{alg:training}
\footnotesize
\SetKwInOut{Input}{Input}
\SetKwInOut{Output}{Init}
\SetAlgoNlRelativeSize{1} 
\SetAlgoNlRelativeSize{0}
\Input{dataset $\mathcal{D}$, pretrained VACE $\bm{F}_{\text{pretrain}}$, pretrained few-step Wan $\bm{W}_{\text{pretrain}}$, Generator $\bm{G}_{\theta}$, Fake model $\bm{F}_{\phi}$, Real model $\bm{F}_{\psi}$, Discriminator $\bm{D}_{\tau}$, learning rates $\eta_1$, $\eta_2$}
\Output{init $\bm{G}_{\theta}$, $\bm{F}_{\phi}$ and $\bm{F}_{\psi}$ with $\bm{F}_{\text{pretrain}}$, init $\bm{G}_{\theta}$ with $\bm{W}_{\text{pretrain}}$.}

\For{$\text{step} \gets 0$ \textbf{to} $\text{max\_step}$}{
    $\triangleright$ \algohighlight{\textbf{Update Generator $\bm{G}_{\theta}$}}: \\
    Sample $\bm{z} \sim \mathcal{N}(\bm{0}, \bm{I})$, $\bm{\hat{x}}_0 \gets \bm{G}_{\theta}(\bm{z})$ \\
    Sample timestamp $t$, $\bm{x}_t \gets \text{add\_noise}(\bm{\hat{x}}_0, t)$ \\
    if $\text{step} \% 2 = 0$: \\
    \quad Compute $\mathcal{L}_{\theta} = \mathcal{L}_{\text{OTD}}(\theta) + \lambda \mathcal{L}_{\text{DMD}}(\theta)$ via~\eqref{eqn:otd} and~\eqref{eqn:dmd} \\
    else: \\ 
    \quad Compute $\mathcal{L}_{\theta} = \mathcal{L}_{\text{GAN}}(\theta)$ via~\eqref{eqn:gan1} \\
    
    Update $\theta \gets \theta - \eta_1 \nabla_{\theta} \mathcal{L}_{\theta}$ \\

    $\triangleright$ \algohighlight{\textbf{Update Fake model $\bm{F}_{\phi}$ and Discriminator $\bm{D}_{\tau}$}}: \\
    Sample $\bm{z} \sim \mathcal{N}(\bm{0}, \bm{I})$, $\bm{\hat{x}}_0 \gets \bm{G}_{\theta}(\bm{z})$, $x^{\text{real}} \sim \mathcal{D}$ \\
    Sample timestamp $t$, $\bm{x}_t^{\text{fake}}/\bm{x}_t^{\text{real}} \gets \text{add\_noise}(\bm{\hat{x}}_0/\bm{x}^{\text{real}}, t)$ \\

    if $\text{step} \% 2 = 0$: \\
    \quad Compute diffusion denoising loss $\mathcal{L}_{\text{Denoising}}(\phi)$ \\
    \quad Update $\phi \gets \phi - \eta_2 \nabla_{\phi} \mathcal{L}_{\text{Denoising}}$ \\
    else: \\ 
    \quad Compute $\mathcal{L}_{\tau} = \mathcal{L}_{\text{GAN}}(\tau)$ via~\eqref{eqn:gan2} \\
    \quad Update $\tau \gets \tau - \eta_2 \nabla_{\tau} \mathcal{L}_{\tau}$ \\
}
\end{algorithm}
\vspace{-10pt}

\section{Dataset}
\label{sec:dataset}

\begin{figure*}[t]
    \centering
    \includegraphics[width=1.0\textwidth]{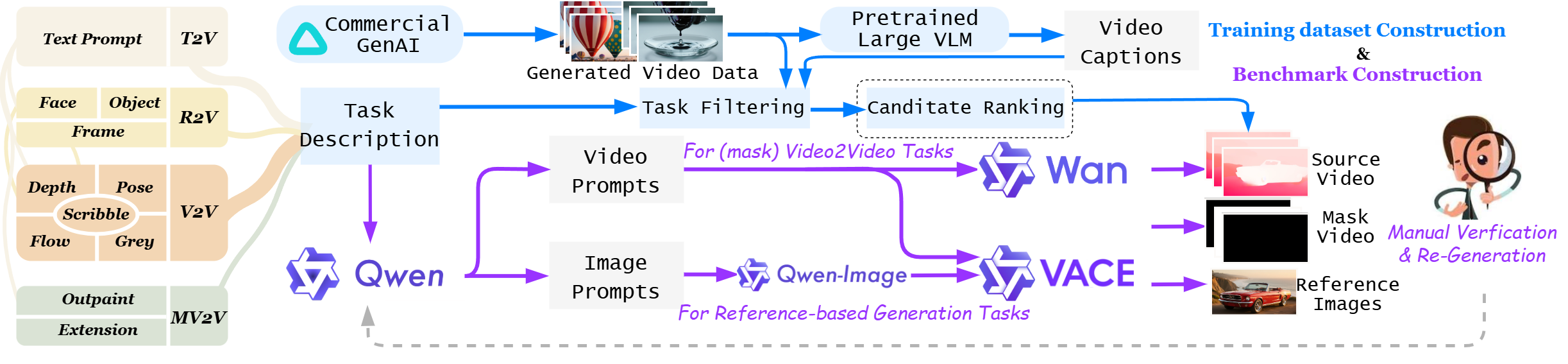}
    \caption{\textbf{Construction pipeline of Training dataset (\textcolor[RGB]{2,127,255}{the blue arrow}) and \methodbench (\textcolor[RGB]{153,51,255}{the purple arrow}).}
    }
    \label{fig:benchmark1}
\end{figure*}

\subsection{Dataset Construction} 
We construct the training dataset fully automatically. 
We generate 250,000 4K videos from Artgrid~\cite{artgrid} spanning diverse content types, and rescale them to $832\times 480$ for training. 
For each video, we use InternVL~\cite{chen2024internvl} to generate a caption, which serves both as a text prompt for generation and as a criterion for filtering. 
As an example, for the \emph{pose}-conditioned data pipeline, we first exclude videos without human subjects by applying Qwen3~\cite{yang2025qwen3} to the video captions. 
We then sample batches of videos and their captions from this filtered pool, derive pose videos from the selected videos, and compute a score defined as the weighted average of inter-frame keypoint distances. 
Finally, we sort videos by this score in descending order and retain the top-ranked pose videos and their captions as training data for the pose task. 
We apply analogous task-specific procedures for the remaining tasks, yielding a unified multi-task video dataset suitable for training across all creation settings. 
Per-task training set sizes are reported in the Appendix.

\subsection{UVCBench}
Video generation has advanced rapidly. 
VBench~\cite{huang2024vbench} and VBench++~\cite{huang2024vbench++} provide established benchmarks for text-to-video and image-to-video tasks. 
However, a comprehensive and standardized benchmark for the arising video creation tasks is still lacking. 
Recently, VACE proposed the VACE-Benchmark with 240 high-quality videos across 12 creation tasks, but it lacks evaluation samples for composite tasks. 
To address this gap and advance this field, we curate \methodbench, a new open-source comprehensive benchmark for unified video creation. 
\methodbench covers 18 tasks—8 single-condition and 10 composite-condition settings. 
Single-condition tasks are defined by the input conditioning signal, including pose, depth, optical flow, greyscale, scribble, reference image (face or object), canvas outpainting, and temporal extension (first, last, or random clip). 
Composite tasks combine either a reference image or a first-frame image with a second, video-based condition (pose, depth, optical flow, greyscale, or scribble). 

We construct the benchmark via an automated pipeline and generate 20 videos per task. 
As shown in Figure~\ref{fig:benchmark1}, we first elaborate task descriptions and prompt Qwen3-Max~\cite{yang2025qwen3} to produce candidate video prompts; for reference-based tasks, this yields paired image–video prompts. 
For V2V and MV2V tasks, we synthesize high-resolution exemplar videos using Wan2.1-14B. 
For R2V and composite tasks, we first create reference images with Qwen-Image~\cite{wu2025qwenimagetechnicalreport} from image prompts, then use VACE-14B to generate exemplar videos conditioned on the reference images and video prompts.
After exemplar generation, an annotator derives the corresponding source inputs and mask videos required by each task. 
Finally, the authors manually verify the results and regenerate failure cases.

\section{Experiments}
\label{sec:experiments}

\subsection{Experimental Setup}

\noindent\textbf{Implementation Details.} 
\method is trained based on Wan2.1-VACE-14B~\cite{jiang2025vace}. 
The training consists of two stages. 
\textit{Stage 1} follows the Self-Forcing pipeline~\cite{huang2025self} to distill Wan2.1–T2V–14B~\cite{wan2025wan} into a few-step generator. 
We train using only video captions from our Artgrid dataset for 1,500 steps. 
The learning rates for the generator and the critic are $2\times 10^{-6}$ and $4\times 10^{-7}$, respectively. 
The TTUR ratio is 5. 
\textit{Stage 2} initializes the generator from Wan2.1–VACE–14B, further pre-initialized with the Stage-1 few-step Wan2.1–T2V–14B weights. 
This stage uses multi-task video data comprising 8 single-task and 10 composite-task settings and trains for 1,200 steps. 
The learning rates for the generator and the critic are $1\times 10^{-6}$ and $4\times 10^{-7}$, respectively. 
The TTUR ratio is 5. 
Both stages use Adam optimizer~\cite{adam2014method}. 
All experiments are conducted on 4 NVIDIA H200 GPUs with a batch size of 1 per GPU, and we adopt gradient checkpointing with a size of 4 to optimize memory usage during training.

\begin{table*}[t]
\caption{
\textbf{Quantitative comparison for various methods on \methodbench}. We bold the \textbf{best results} and underline the \underline{second-best results}. 
}

\centering
\footnotesize
\setlength{\tabcolsep}{2.5pt}
\begin{tabular}{l|l|c|c|ccccccc|cccc}

\toprule

Type & Method & Base Model & \#NFE$\downarrow$ & \multicolumn{7}{c|}{\textbf{Video Quality \& Video Consistency}} 
& \multicolumn{4}{c}{\textbf{User Study}} \\

& & & & \rotatebox{90}{\begin{minipage}{1.8cm} {\tikz\fill[video_quality] (0,0) circle (.5ex);} \centering Aesthetic Quality\end{minipage}}
& \rotatebox{90}{\begin{minipage}{1.8cm} {\tikz\fill[video_quality] (0,0) circle (.5ex);} \centering Background Consistency\end{minipage}}
& \rotatebox{90}{\begin{minipage}{1.8cm} {\tikz\fill[video_quality] (0,0) circle (.5ex);} \centering Dynamic Degree\end{minipage}}
& \rotatebox{90}{\begin{minipage}{1.8cm} {\tikz\fill[video_quality] (0,0) circle (.5ex);} \centering Imaging Quality\end{minipage}}
& \rotatebox{90}{\begin{minipage}{1.8cm} {\tikz\fill[video_quality] (0,0) circle (.5ex);} \centering Motion Smoothness\end{minipage}}
& \rotatebox{90}{\begin{minipage}{1.8cm} {\tikz\fill[video_quality] (0,0) circle (.5ex);} \centering Subject Consistency\end{minipage}}
& \rotatebox{90}{\begin{minipage}{1.8cm} {\tikz\fill[score_average] (0,0) circle (.5ex);} \centering Normalized Average\end{minipage}}
& \rotatebox{90}{\begin{minipage}{1.8cm} {\tikz\fill[user_study] (0,0) circle (.5ex);} \centering Prompt Following\end{minipage}}
& \rotatebox{90}{\begin{minipage}{1.8cm} {\tikz\fill[user_study] (0,0) circle (.5ex);} \centering Temporal Consistency\end{minipage}} 
& \rotatebox{90}{\begin{minipage}{1.8cm} {\tikz\fill[user_study] (0,0) circle (.5ex);} \centering Video Quality\end{minipage}}
& \rotatebox{90}{\begin{minipage}{1.8cm} {\tikz\fill[score_average] (0,0) circle (.5ex);} \centering Average\end{minipage}} \\

\midrule 

\multirow{5}{*}{Depth}
& Control-A-Video~\cite{chen2023control}  &SD-1.5  &100 &57.22\%  &92.89\%  &20.00\%  &66.95\%  &98.06\%  &91.85\%  &71.16\%  &2.04  &1.85  &1.12  &1.67 \\ 
& ControlVideo~\cite{zhang2023controlvideo}  &SD-1.5  &100  &\textbf{64.50\%}  &97.99\%  &5.00\%  &\textbf{71.76\%}  &98.10\%  &97.41\%  &72.46\%  &2.84  &2.78  &2.31  &2.64 \\ 
& VACE~\cite{jiang2025vace} &LTX-0.9B  &80  &58.62\%  &\textbf{98.18\%}  &30.00\%  &70.14\%  &\underline{99.29\%}  &\underline{97.81\%}  &75.67\%  &2.89  &2.82  &2.24  &2.65 \\
& VACE~\cite{jiang2025vace} &Wan-14B  &100  &\underline{64.36\%}  &97.63\%  &\underline{35.00\%}  &70.29\%  &99.17\%  &97.58\%  &\underline{77.34\%}  &\underline{4.33}  &\textbf{4.41}  &\textbf{4.65}  &\textbf{4.46} \\
& \cellcolor{tabhighlight}\methodbf (Ours)  &\cellcolor{tabhighlight}Wan-14B  &\cellcolor{tabhighlight}4  &\cellcolor{tabhighlight}64.28\%  &\cellcolor{tabhighlight}\textbf{98.18\%}  &\cellcolor{tabhighlight}\textbf{40.00\%}  &\cellcolor{tabhighlight}\underline{71.24\%}  &\cellcolor{tabhighlight}\textbf{99.35\%}  &\cellcolor{tabhighlight}\textbf{97.96\%}  &\cellcolor{tabhighlight}\textbf{78.50\%}  &\cellcolor{tabhighlight}\textbf{4.51}  &\cellcolor{tabhighlight}\underline{4.27}  &\cellcolor{tabhighlight}\underline{4.60}  &\cellcolor{tabhighlight}\textbf{4.46} \\
\midrule 

\multirow{5}{*}{Pose} 
& ControlVideo~\cite{zhang2023controlvideo} &SD-1.5  &100  &63.32\%  &\textbf{96.82\%}  &10.00\%  &\underline{69.88\%}  &98.50\%  &94.45\%  &72.16\%  &2.79  &2.65  &2.18  &2.54 \\
& Follow-Your-Pose~\cite{ma2024follow} &SD-1.4  &50  &50.36\%  &88.61\%  &40.00\%  &68.73\%  &91.78\%  &77.81\%  &69.55\%  &1.63  &1.42  &1.05  &1.37 \\ 
& VACE~\cite{jiang2025vace} &LTX-0.9B  &80  &59.72\%  &95.67\%  &45.00\%  &69.12\%  &98.10\%  &\textbf{94.80\%}  &77.06\%  &2.71  &2.72  &2.27  &2.57 \\
& VACE~\cite{jiang2025vace} &Wan-14B  &100  &\textbf{64.99\% } &95.48\%  &\underline{55.00\%}  &69.15\%  &\underline{98.64\%}  &94.07\%  &\underline{79.56\%}  &\underline{4.44}  &\textbf{4.50}  &\textbf{4.36}  &\underline{4.43} \\
& \cellcolor{tabhighlight}\methodbf (Ours)  &\cellcolor{tabhighlight}Wan-14B  &\cellcolor{tabhighlight}4  &\cellcolor{tabhighlight}\underline{63.50\%}  &\cellcolor{tabhighlight}\underline{95.88\%}  &\cellcolor{tabhighlight}\textbf{60.00\%}  &\cellcolor{tabhighlight}\textbf{70.65\%}  &\cellcolor{tabhighlight}\textbf{98.69\%}  &\cellcolor{tabhighlight}\underline{94.53\%}  &\cellcolor{tabhighlight}\textbf{80.54\%}  &\cellcolor{tabhighlight}\textbf{4.75}  &\cellcolor{tabhighlight}\underline{4.37}  &\cellcolor{tabhighlight}\underline{4.28}  &\cellcolor{tabhighlight}\textbf{4.47} \\
\midrule 

\multirow{4}{*}{Flow} 
& FLATTEN~\cite{cong2023flatten} &SD-2.1  &100  &\textbf{63.86\%}  &\textbf{96.68\%}  &60.00\%  &53.28\%  &96.35\%  &\underline{93.39\%}  &77.26\%  &3.05  &2.89  &3.31  &3.08 \\ 
& VACE~\cite{jiang2025vace} &LTX-0.9B  &80  &55.47\%  &95.97\%  &65.00\%  &57.51\%  &97.62\%  &92.90\%  &77.41\%  &2.56  &2.43  &2.93  &2.64 \\
& VACE~\cite{jiang2025vace} &Wan-14B  &100  &\underline{62.78\%}  &95.78\%  &\textbf{75.00\%}  &\underline{58.01\%}  &\underline{97.77\%}  &92.75\%  &\textbf{80.35\%}  &\underline{4.39}  &\underline{4.40}  &\underline{4.55}  &\underline{4.45} \\
& \cellcolor{tabhighlight}\methodbf (Ours)  &\cellcolor{tabhighlight}Wan-14B  &\cellcolor{tabhighlight}4  &\cellcolor{tabhighlight}\underline{62.78\%}  &\cellcolor{tabhighlight}\underline{96.36\%}  &\cellcolor{tabhighlight}\underline{70.00\%}  &\cellcolor{tabhighlight}\textbf{59.22\%}  &\cellcolor{tabhighlight}\textbf{98.84\% } &\cellcolor{tabhighlight}\textbf{93.93\%}  &\cellcolor{tabhighlight}\underline{80.18\%}  &\cellcolor{tabhighlight}\textbf{4.45}  &\cellcolor{tabhighlight}\textbf{4.52}  &\cellcolor{tabhighlight}\textbf{4.57}  &\cellcolor{tabhighlight}\textbf{4.51} \\
\midrule 

\multirow{4}{*}{Scribble} 
& ControlVideo~\cite{zhang2023controlvideo} &SD-1.5  &100  &\underline{54.79\%}  &96.19\%  &5.00\%  &\underline{64.30\%}  &98.53\%  &\underline{94.64\%}  &68.91\%  &3.78  &2.27  &2.39  &2.81 \\
& VACE~\cite{jiang2025vace} &LTX-0.9B  &80  &44.69\%  &96.88\%  &40.00\%  &61.06\%  &\textbf{99.17\%}  &93.51\%  &72.55\%  &3.99  &3.71  &3.46  &3.72 \\
& VACE~\cite{jiang2025vace} &Wan-14B  &100  &\textbf{55.70\%}  &\underline{96.91\%}  &\textbf{55.00\%}  &58.69\%  &\underline{98.99\%}  &94.51\%  &\underline{76.63\%}  &\textbf{4.67}  &\underline{4.67}  &\textbf{4.82}  &\textbf{4.72} \\
& \cellcolor{tabhighlight}\methodbf (Ours)  &\cellcolor{tabhighlight}Wan-14B  &\cellcolor{tabhighlight}4  &\cellcolor{tabhighlight}54.24\%  &\cellcolor{tabhighlight}\textbf{97.13\%}  &\cellcolor{tabhighlight}\underline{50.00\%}  &\cellcolor{tabhighlight}\textbf{65.19\% } &\cellcolor{tabhighlight}98.90\%  &\cellcolor{tabhighlight}\textbf{95.20\%}  &\cellcolor{tabhighlight}\textbf{76.77\%}  &\cellcolor{tabhighlight}\underline{4.62}  &\cellcolor{tabhighlight}\textbf{4.74}  &\cellcolor{tabhighlight}\underline{4.78}  &\cellcolor{tabhighlight}\underline{4.71} \\
\midrule 

\multirow{3}{*}{Grey} 
& VACE~\cite{jiang2025vace} &LTX-0.9B  &80  &60.64\%  &98.09\%  &5.00\%  &58.29\%  &\textbf{99.37\%}  &97.14\%  &69.75\%  &4.30  &4.07  &3.75  &4.04 \\
& VACE~\cite{jiang2025vace} &Wan-14B  &100  &\textbf{63.76\%}  &\textbf{98.11\% } &\textbf{15.00\%}  &\textbf{60.90\%}  &99.30\%  &\underline{97.56\%}  &\textbf{72.44\%}  &\underline{4.65}  &\textbf{4.79}  &\underline{4.81}  &\underline{4.75} \\
& \cellcolor{tabhighlight}\methodbf (Ours)  &\cellcolor{tabhighlight}Wan-14B  &\cellcolor{tabhighlight}4  &\cellcolor{tabhighlight}\underline{62.13\%}  &\cellcolor{tabhighlight}\textbf{98.11\%}  &\cellcolor{tabhighlight}\textbf{15.00\%}  &\cellcolor{tabhighlight}\underline{60.60\%}  &\cellcolor{tabhighlight}\underline{99.34\%}  &\cellcolor{tabhighlight}\textbf{97.62\%}  &\cellcolor{tabhighlight}\underline{72.13\%}  &\cellcolor{tabhighlight}\textbf{4.70}  &\cellcolor{tabhighlight}4.75  &\cellcolor{tabhighlight}\textbf{4.84}  &\cellcolor{tabhighlight}\textbf{4.76} \\
\midrule 

\multirow{4}{*}{Outpaint} 
& Follow-Your-Canvas~\cite{chen2024follow}  &SD-2.1  &80  &51.98\%  &97.13\%  &\underline{25.00\%}  &66.95\%  &98.92\%  &95.55\%  &72.59\%  &3.75  &3.89  &3.51  &3.72 \\ 
& VACE~\cite{jiang2025vace} &LTX-0.9B  &80  &58.78\%  &98.07\%  &\underline{25.00\%}  &\underline{70.74\%}  &\textbf{99.20\%}  &\underline{97.35\%}  &74.86\%  &4.20  &3.44  &3.77  &3.80 \\
& VACE~\cite{jiang2025vace} &Wan-14B  &100  &\underline{62.66\%}  &\underline{98.13\%}  &\underline{25.00\%}  &70.56\%  &\underline{99.18\%}  &\textbf{97.41\%}  &\underline{75.49\%}  &\textbf{4.61}  &\underline{4.63}  &\underline{4.47}  &\underline{4.57} \\
& \cellcolor{tabhighlight}\methodbf (Ours)  &\cellcolor{tabhighlight}Wan-14B  &\cellcolor{tabhighlight}4  &\cellcolor{tabhighlight}\textbf{63.18\%}  &\cellcolor{tabhighlight}\textbf{98.22\%}  &\cellcolor{tabhighlight}\textbf{30.00\%}  &\cellcolor{tabhighlight}\textbf{71.38\%}  &\cellcolor{tabhighlight}99.16\%  &\cellcolor{tabhighlight}97.34\%  &\cellcolor{tabhighlight}\textbf{76.54\%}  &\cellcolor{tabhighlight}\underline{4.52}  &\cellcolor{tabhighlight}\textbf{4.70}  &\cellcolor{tabhighlight}\textbf{4.63}  &\cellcolor{tabhighlight}\textbf{4.62} \\
\midrule 

\multirow{3}{*}{Extension} 
& VACE~\cite{jiang2025vace} &LTX-0.9B  &80  &51.54\%  &\textbf{96.48\%}  &20.00\%  &62.22\%  &\textbf{99.21\%}  &\textbf{94.67\%}  &70.68\%  &3.50  &2.85  &3.14  &3.16 \\
 & VACE~\cite{jiang2025vace} &Wan-14B  &100  &\underline{57.52\%}  &\underline{95.46\%}  &\underline{55.00\%}  &\underline{63.36\%}  &98.99\%  &\underline{92.27\%}  &\underline{77.10\%}  &\textbf{4.55}  &\textbf{4.54}  &\textbf{4.48}  &\textbf{4.52} \\
& \cellcolor{tabhighlight}\methodbf (Ours)  &\cellcolor{tabhighlight}Wan-14B  &\cellcolor{tabhighlight}4  &\cellcolor{tabhighlight}\textbf{58.93\%}  &\cellcolor{tabhighlight}94.18\%  &\cellcolor{tabhighlight}\textbf{75.00\%}  &\cellcolor{tabhighlight}\textbf{65.11\% } &\cellcolor{tabhighlight}\textbf{99.21\%}  &\cellcolor{tabhighlight}91.09\%  &\cellcolor{tabhighlight}\textbf{80.53\%} &\cellcolor{tabhighlight}\underline{4.32}  &\cellcolor{tabhighlight}\underline{4.40}  &\cellcolor{tabhighlight}\underline{4.35}  &\cellcolor{tabhighlight}\underline{4.36} \\
\midrule 

\multirow{5}{*}{R2V} 
& Keling1.6~\cite{kling} &-  &-  &63.61\%  &93.85\%  &\textbf{90.00\%}  &\underline{64.46\%}  &\textbf{98.95\%}  &\underline{90.11\%}  &\textbf{83.50\%}  &\textbf{4.82}  &\textbf{4.78}  &\textbf{4.85}  &\textbf{4.82} \\
 &Vidu2.0~\cite{vidu} &-  &-  &60.79\%  &93.22\%  &\underline{85.00\%}  &54.26\%  &97.54\%  &87.42\%  &79.71\%  &4.52  &4.33  &4.50  &4.45 \\
 &VACE~\cite{jiang2025vace} &LTX-0.9B  &80  &50.18\%  &91.31\%  &50.00\%  &55.88\%  &\underline{98.84\%}  &85.49\%  &71.95\%  &2.36  &2.03  &3.11  &2.50 \\
 &VACE~\cite{jiang2025vace} &Wan-14B  &100  &\underline{67.39\%}  &\textbf{95.39\%}  &80.00\%  &62.82\%  &97.96\%  &\textbf{91.70\%}  &\underline{82.54\%}  &\underline{4.65}  &\underline{4.71}  &4.63  &\underline{4.66} \\
& \cellcolor{tabhighlight}\methodbf (Ours)  &\cellcolor{tabhighlight}Wan-14B  &\cellcolor{tabhighlight}4  &\cellcolor{tabhighlight}\textbf{69.70\%}  &\cellcolor{tabhighlight}\underline{94.34\%}  &\cellcolor{tabhighlight}70.00\%  &\cellcolor{tabhighlight}\textbf{65.93\%}  &\cellcolor{tabhighlight}98.15\%  &\cellcolor{tabhighlight}89.88\%  &\cellcolor{tabhighlight}81.32\%  &\cellcolor{tabhighlight}4.62  &\cellcolor{tabhighlight}4.65  &\cellcolor{tabhighlight}\underline{4.66}  &\cellcolor{tabhighlight}4.64 \\

\bottomrule
\end{tabular}
\label{tab:comp1}
\vspace{5pt}
\end{table*}

\begin{table*}[t]
\caption{
\textbf{Ablation Study on \methodbench}. Base Model is VACE-Wan2.1-14B. AccWanInit refers to initializing the Wan blocks in the VACE model using a few-step Wan model.
}
\centering
\small
\setlength{\tabcolsep}{3.8pt}
\begin{tabular}{l|cccc|cccccccc}
\toprule
&\textbf{DMD} & \textbf{OTD} & \textbf{GAN} & \textbf{AccWanInit} & \textbf{Depth} & \textbf{Pose} & \textbf{Flow} & \textbf{Scribble} & \textbf{Grey} & \textbf{Outpaint} & \textbf{Extention} & \textbf{R2V} \\ \midrule 
\textbf{(1)}&$\usym{2717}$ & $\usym{2717}$ &$\usym{2717}$ &$\usym{2713}$  &77.82\%  &80.07\%  &79.71\%  &75.17\%  &71.69\%  &75.38\%  &76.01\%  &80.17\%    \\ \hline
\textbf{(2)}&$\usym{2713}$ & $\usym{2717}$ &$\usym{2717}$ &$\usym{2717}$  & 76.89\%  & 78.34\% & 79.12\% & 75.79\% & 71.70\% & 74.21\% & 77.33\% & 76.45\%   \\ \hline
\textbf{(3)}&$\usym{2713}$ & $\usym{2713}$ &$\usym{2717}$ &$\usym{2713}$  & 78.24\%  & 80.14\% & 79.95\% & 76.90\% & 71.59\% & 76.44\% & 79.01\% & 77.66\%   \\ \hline
\textbf{(4)}&$\usym{2713}$ & $\usym{2717}$ &$\usym{2713}$ &$\usym{2713}$  & 78.05\%  & \textbf{80.79\%} & 80.15\% & 76.58\% & 71.98\% & 76.40\% & 78.88\% & 78.40\%   \\ \hline
\textbf{(5)}&$\usym{2713}$ & $\usym{2713}$ &$\usym{2713}$ &$\usym{2717}$  &77.15\%   &79.83\%   &79.16\%   &\textbf{77.19\%}    &\textbf{72.21\%}   &75.57\%   &79.76\%   &77.00\%    \\ \hline 
\methodbf&$\cellcolor{tabhighlight} \usym{2713}$ &\cellcolor{tabhighlight} $\usym{2713}$ &\cellcolor{tabhighlight} $\usym{2713}$ &\cellcolor{tabhighlight} $\usym{2713}$  &\cellcolor{tabhighlight}\textbf{78.50\%}   &\cellcolor{tabhighlight}80.54\% &\cellcolor{tabhighlight}\textbf{80.18\%}  &\cellcolor{tabhighlight}76.77\%  &\cellcolor{tabhighlight}72.13\%  &\cellcolor{tabhighlight}\textbf{76.54\%}  &\cellcolor{tabhighlight}\textbf{80.53\%}  &\cellcolor{tabhighlight}\textbf{81.32\%}    \\ \bottomrule
\end{tabular}
\label{tab:aba1}
\vspace{5pt}
\end{table*}

\noindent\textbf{Baselines.} 
We evaluate \method against VACE~\cite{jiang2025vace}, the only open-source all-in-one video creation model, and further include task-specific online and offline methods for comparison. 
(1) \emph{Video-to-video (V2V)}, under \emph{depth} conditioning—Control-A-Video~\cite{chen2023control} and ControlVideo~\cite{zhang2023controlvideo}; under \emph{pose} conditioning—ControlVideo and Follow-Your-Pose~\cite{ma2024follow}; under \emph{scribble} conditioning—ControlVideo; under \emph{optical-flow} conditioning—FLATTEN~\cite{cong2023flatten}. 
(2) \emph{MV2V}: for \emph{outpainting}—Follow-Your-Canvas~\cite{chen2024follow}. 
(3) \emph{Reference-based video generation (R2V)}: online systems Keling-1.6~\cite{kling} and Vidu-2.0~\cite{vidu}. 

\begin{figure*}[t]
    \centering
    \includegraphics[width=1.0\textwidth]{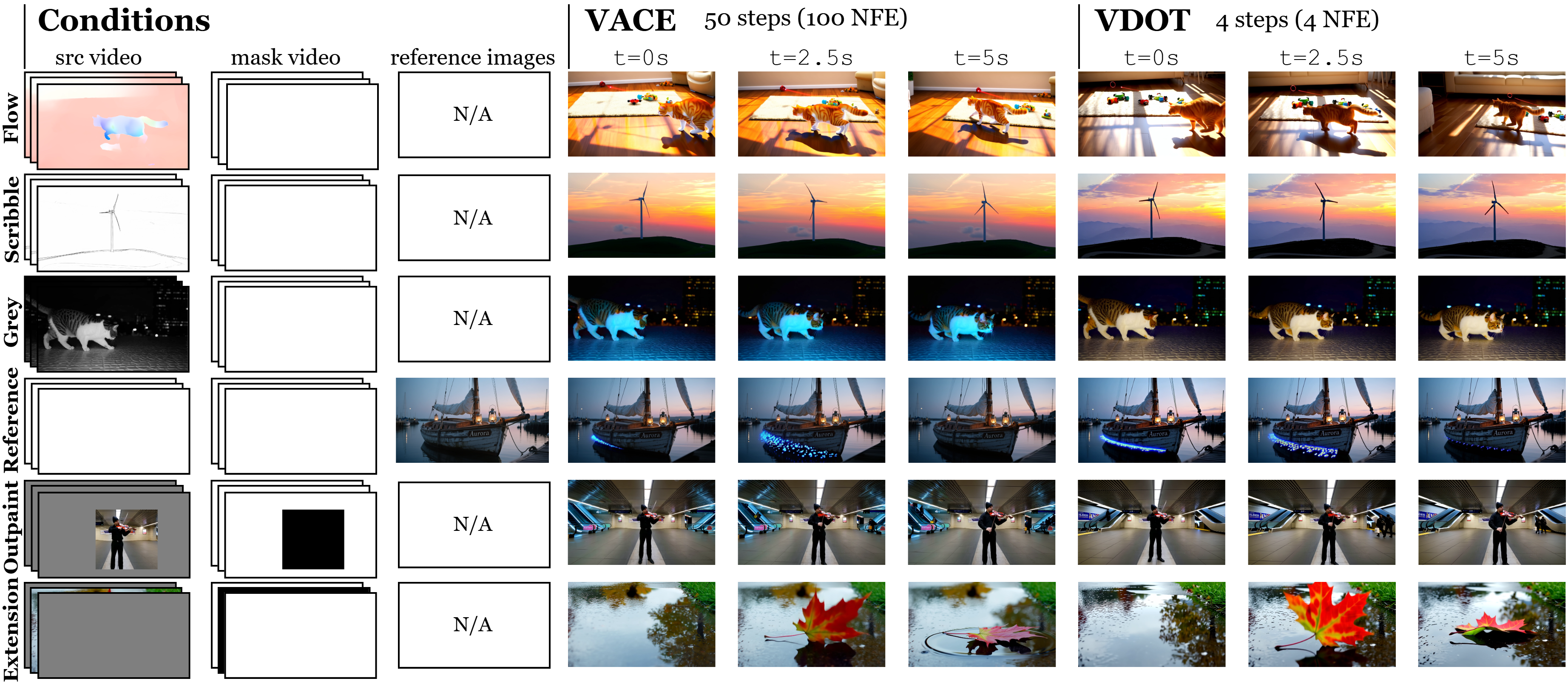}
    \caption{\textbf{Qualitative comparison between generated videos.} \method can generate comparable visual fidelity with 4 denoising steps compared with VACE-Wan2.1-14B with 50 denoising steps. Refer to the Appendix for more visualizations. Zoom in for more details.
    }
    \label{fig:visualization1}
\end{figure*}

\begin{figure}[t]
    \includegraphics[width=0.45\textwidth]{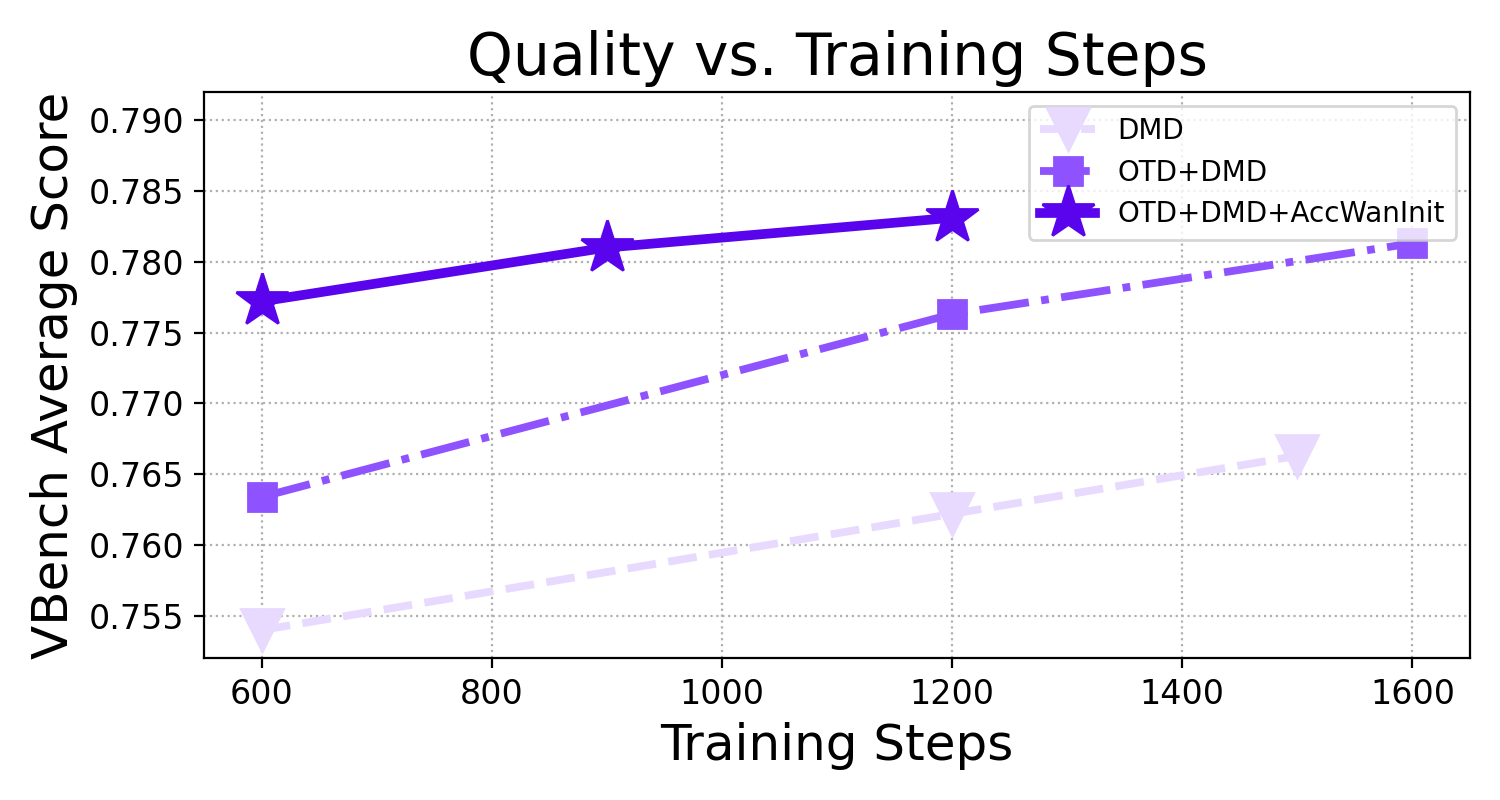}
    \vspace{-3pt}
    \caption{\textbf{Training efficiency comparison.} 
    }
    \label{fig:efficiency}
\end{figure}

\noindent\textbf{Evaluation.} 
For a comprehensive benchmarking, we use our proposed \methodbench for evaluation. 
We use VBench~\cite{huang2024vbench} to assess video quality and consistency of generated results across six dimensions: Aesthetic Quality, Background Consistency, Dynamic Degree, Imaging Quality, Motion Smoothness, and Subject Consistency. 
In addition to automatic scoring, we conduct a user study for subjective evaluation. 
Following VACE, annotators rate Prompt Following, Temporal Consistency, and Overall Video Quality. 
Scores are reported on a 1–5 Likert scale. 

\subsection{Quantitative and Qualitative Comparisons}

Table~\ref{tab:comp1} presents comprehensive quantitative comparisons across multiple video creation tasks on \methodbench, covering both objective video metrics and subjective user preference. 
In terms of video quality and temporal consistency, \method delivers competitive or superior performance while requiring drastically fewer inference steps. 
In particular, for \emph{Imaging Quality}, \method attains the best or second-best scores across all tasks. 
These results highlight the efficiency and generality of \method. 
We also include a user study comparison to assess \emph{Prompt Following}, \emph{Temporal Consistency}, and \emph{Overall Video Quality} across all creation tasks. 
We establish a website and invite 20 volunteers to provide ratings. 
For each metric, scores are averaged across raters. 
In summary, \method achieves encouraging performance compared with the open-source baselines or commercial products in average user preference, validating the effectiveness of our proposed framework. 

Figure~\ref{fig:visualization1} presents qualitative comparisons between VACE and \method on five tasks—\emph{flow}, \emph{scribble}, \emph{grey}, \emph{outpainting}, and \emph{temporal extension}. 
\method achieves comparable visual appearance and structural consistency to VACE while using far fewer inference steps. 
More quantitative results and visualizations about the composite tasks can be found in the Appendix.

\subsection{Ablation Study}

Table~\ref{tab:aba1} presents a comprehensive ablation study evaluating the contributions of each component in \method using VACE-Wan2.1-14B as the base model. 
All variants (except row 1) are trained for 1{,}200 steps. 
Row (1) is the baseline, where we replace only the Wan blocks in VACE~\cite{jiang2025vace} with a distilled Wan. 
Row (2) corresponds to the Self-Forcing~\cite{huang2025self} paradigm and employs only the DMD loss. 
Removing GAN objectives (row 3) or OTD objective (row 4) leads to consistent degradation across most metrics, confirming their necessary roles in video distillation. 
Row (5) indicates that AccWanInit supplies a stronger initialization, substantially improving training efficiency. 
As shown in Figure~\ref{fig:efficiency}, we also plot a quality versus training steps graph to demonstrate the training efficiency gains afforded by OTD and AccWanInit.
\vspace{4pt}
\section{Conclusion}
\label{sec:conclusion}

In this work, we propose \method, an efficient unified video creation model. 
We introduce a novel computational optimal transport–based distribution matching objective which, together with DMD and GAN losses, enables few-step distillation of video models with improved visual quality and enhanced both training and inference efficiency. 
To support multi-task video creation, we develop a fully automated pipeline for training data annotation and filtering, and curate a unified benchmark, \methodbench, to enable fair and comprehensive evaluation. 
Experiments on \methodbench show that our method achieves encouraging performance in unified video creation using only few denoising steps.

\clearpage
{
    \small
    \bibliographystyle{ieeenat_fullname}
    \bibliography{main}
}

\clearpage
\setcounter{page}{1}
\maketitlesupplementary

\section{Implementation Details} 

\subsection{Dataset Composition}

\begin{table}[t]
\caption{
\textbf{Overview of the training dataset composition.}
}
\centering
\small
\setlength{\tabcolsep}{4.5pt}
\begin{tabular}{cll|c}
\toprule
\multicolumn{2}{l}{\emph{Task}}                               & \texttt{Input} & \#Examples \\ \midrule
\multicolumn{1}{c|}{\multirow{5}{*}{\emph{V2V}}}  & \emph{Depth}     &\texttt{txt+video}       & $\mathcal{O}(6)$K           \\ \cline{2-4} 
\multicolumn{1}{c|}{}                      & \emph{Pose}      &\texttt{txt+video}       &$\mathcal{O}(12)$K             \\ \cline{2-4} 
\multicolumn{1}{c|}{}                      & \emph{Grey} &\texttt{txt+video}       &$\mathcal{O}(6)$K             \\ \cline{2-4} 
\multicolumn{1}{c|}{}                      & \emph{Scribble}  &\texttt{txt+video}       &$\mathcal{O}(6)$K             \\ \cline{2-4} 
\multicolumn{1}{c|}{}                      & \emph{Flow}      &\texttt{txt+video}       &$\mathcal{O}(6)$K             \\ \hline
\multicolumn{1}{c|}{\multirow{2}{*}{\emph{MV2V}}} & \emph{Outpaint}  &\texttt{txt+video+mask}       &$\mathcal{O}(12)$K             \\ \cline{2-4} 
\multicolumn{1}{c|}{}                      & \emph{Extension} &\texttt{txt+video+mask}       &$\mathcal{O}(12)$K             \\ \hline
\multicolumn{1}{c|}{\emph{R2V}}                   & \emph{Reference} &\texttt{txt+image}       &$\mathcal{O}(15)$K             \\ \hline
\multicolumn{2}{l}{\emph{Composite Tasks}}                    &\texttt{txt+video+mask+image}       &$\mathcal{O}(20)$K             \\ \bottomrule
\end{tabular}
\label{tab:training}
\end{table}

Table~\ref{tab:training} details the composition of our training dataset across different tasks.
For the V2V tasks, we curate datasets for five distinct condition signals: \emph{depth}, \emph{pose}, \emph{greyscale}, \emph{scribble}, and \emph{flow}. 
Each is formulated as a conditional video-to-video generation task containing approximately 6-12K samples. 
The MV2V tasks extend this framework by incorporating mask inputs to support \emph{outpainting} and \emph{temporal extension}, with roughly 12K samples per task. 
For the R2V task, we include 15K examples where generation is conditioned jointly on text prompts and reference images. 
Finally, for the composite tasks, we implement dual control signals: (1) a first-frame or reference image, and (2) one of the five V2V modalities. 
This results in 10 distinct subtasks, each with 2K samples, totaling 20K composite training examples.

\subsection{Diffusion denoising objective}
When updating the fake model $\bm{F}_{\phi}$, we adopt the standard diffusion denoising loss~\cite{vincent2011connection,ho2020denoising}: 
\begin{equation}
    \mathcal{L}_{\text{Denoising}}(\phi)=\| \bm{F}_{\phi}(\bm{x}_t^{\text{fake}},t) - \bm{\hat{x}}_0 \|^2_2,
\end{equation}
where $\bm{x}_t^{\text{fake}}$ represents the noisy sample obtained by adding noise to the generator output $\bm{\hat{x}}_0$. 

\subsection{Sinkhorn algorithm}
\newcommand{\gray}[1]{\textcolor{gray}{#1}}

\begin{algorithm}[t]
\small
\caption{Log-domain Sinkhorn Algorithm for Entropic Optimal Transport (Equation~\eqref{eqn:eot})}
\label{alg:sinkhorn}

\DontPrintSemicolon
\SetKwInOut{KwIn}{Input}
\SetKwInOut{KwOut}{Output}
\IncMargin{0.5em}

\KwIn{
  Cost matrix $\bm{D} \in \mathbb{R}^{I \times J}$;\\
  Marginals $\bm{u} \in \mathbb{R}^{I}_+$, $\bm{\mu} \in \mathbb{R}^{J}_+$;\\
  Regularization parameter $\epsilon > 0$;\\
  Tolerance $\texttt{tol}$ and iterations $\texttt{max\_iter}$.
}
\KwOut{
  Optimal transport plan $\bm{T}^{*} \in \mathbb{R}^{I \times J}_+$.
}

\tcp{\gray{Precompute log-kernel}}
$\log \bm{K} \gets -\bm{D}/\epsilon$\;

\tcp{\gray{Initialize dual variables}}
$\bm{f}^{(0)} \gets \bm{0}_I$, \quad $\bm{g}^{(0)} \gets \bm{0}_J$\;

\For{$t = 1$ \KwTo $\texttt{max\_iter}$}{
  
  $\bm{f}_{\mathrm{old}} \gets \bm{f}^{(t-1)}$,\quad
  $\bm{g}_{\mathrm{old}} \gets \bm{g}^{(t-1)}$\;

  \tcp{\gray{Update $\bm{f}$}}
  \For{$i = 1$ \KwTo $I$}{
    $s \gets \log \sum_{j=1}^{J} 
      \exp\!\left( \log K_{ij} + g^{(t-1)}_j \right)$\;
    $f^{(t)}_i \gets \log u_i - s$\;
  }

  \tcp{\gray{Update $\bm{g}$}}
  \For{$j = 1$ \KwTo $J$}{
    $s \gets \log \sum_{i=1}^{I}
      \exp\!\left( \log K_{ij} + f^{(t)}_i \right)$\;
    $g^{(t)}_j \gets \log \mu_j - s$\;
  }

  \tcp{\gray{Stopping criterion}}
  \If{$\|\bm{f}^{(t)} - \bm{f}_{\mathrm{old}}\|_1 
    + \|\bm{g}^{(t)} - \bm{g}_{\mathrm{old}}\|_1 < \texttt{tol}$}{
    \textbf{break}\;
  }
}

\tcp{\gray{Recover optimal transport plan}}
\For{$i = 1$ \KwTo $I$}{
  \For{$j = 1$ \KwTo $J$}{
    $\log T^{*}_{ij} \gets f^{(t)}_i + \log K_{ij} + g^{(t)}_j$\;
    $T^{*}_{ij} \gets \exp(\log T^{*}_{ij})$\;
  }
}

\Return $\bm{T}^{*}$\;

\end{algorithm}
 
As illustrated in Algorithm~\ref{alg:sinkhorn}, we solve the entropic optimal transport problem in Equation~\eqref{eqn:eot} using the log-domain Sinkhorn algorithm~\cite{peyre2019computational}. 
Unlike the standard Sinkhorn iterations, the log-domain formulation stabilizes the dual updates by performing all computations in logarithmic space, thereby preventing underflow of the Gibbs kernel and ensuring numerically reliable convergence even with small regularization $\epsilon$.

\section{Quantitative and Qualitative Analysis}

\begin{table*}[t]
\caption{
\textbf{Quantitative evaluations of composite tasks on \methodbench}. We compare the automated score metrics of our \method with VACE on the dimensions of video quality and video consistency.
}

\centering
\footnotesize
\setlength{\tabcolsep}{4.5pt}
\begin{tabular}{l|l|l|c|c|cccccc|c}

\toprule

Condition1 & Condition2 & Method & Base Model & \#NFE$\downarrow$ & \multicolumn{7}{c}{\textbf{Video Quality \& Video Consistency}} \\

& & & & & \rotatebox{90}{\begin{minipage}{1.8cm} {\tikz\fill[video_quality] (0,0) circle (.5ex);} \centering Aesthetic Quality\end{minipage}}
& \rotatebox{90}{\begin{minipage}{1.8cm} {\tikz\fill[video_quality] (0,0) circle (.5ex);} \centering Background Consistency\end{minipage}}
& \rotatebox{90}{\begin{minipage}{1.8cm} {\tikz\fill[video_quality] (0,0) circle (.5ex);} \centering Dynamic Degree\end{minipage}}
& \rotatebox{90}{\begin{minipage}{1.8cm} {\tikz\fill[video_quality] (0,0) circle (.5ex);} \centering Imaging Quality\end{minipage}}
& \rotatebox{90}{\begin{minipage}{1.8cm} {\tikz\fill[video_quality] (0,0) circle (.5ex);} \centering Motion Smoothness\end{minipage}}
& \rotatebox{90}{\begin{minipage}{1.8cm} {\tikz\fill[video_quality] (0,0) circle (.5ex);} \centering Subject Consistency\end{minipage}}
& \rotatebox{90}{\begin{minipage}{1.8cm} {\tikz\fill[score_average] (0,0) circle (.5ex);} \centering Normalized Average\end{minipage}} \\

\midrule 

\multirow{15}{*}{FirstFrame}& \multirow{3}{*}{Depth} 
& VACE~\cite{jiang2025vace} &LTX-0.9B  &80  &64.78\%  &\textbf{96.90\%}  &40.00\%  &66.19\%  &98.96\%  &94.49\%  &76.89\%  \\
& & VACE~\cite{jiang2025vace} &Wan-14B  &100  &\textbf{68.28\%}  &96.74\%  &40.00\%  &68.43\%  &98.92\%  &94.38\%  &77.79\%  \\
& &\cellcolor{tabhighlight}\methodbf (Ours) & \cellcolor{tabhighlight}Wan-14B  & \cellcolor{tabhighlight}4  &\cellcolor{tabhighlight}67.34\% &\cellcolor{tabhighlight}96.67\% &\cellcolor{tabhighlight}\textbf{45.00\%} &\cellcolor{tabhighlight}\textbf{70.09\%} &\cellcolor{tabhighlight}\textbf{98.97\%} &\cellcolor{tabhighlight}\textbf{94.64\%} &\cellcolor{tabhighlight}\textbf{78.78\%}  \\

\cline{2-12} 

& \multirow{3}{*}{Pose} 
& VACE~\cite{jiang2025vace} &LTX-0.9B  &80  &64.75\%  &\textbf{96.29\%}  &15.00\%  &63.90\%  &\textbf{99.13\%}  &94.63\%  &72.28\%  \\
& & VACE~\cite{jiang2025vace} &Wan-14B  &100  &\textbf{66.66\%}  &95.91\%  &\textbf{40.00\%}  &66.99\%  &98.94\%  &\textbf{94.70\%}  &\textbf{77.19\%}   \\
& &\cellcolor{tabhighlight}\methodbf (Ours) & \cellcolor{tabhighlight}Wan-14B  & \cellcolor{tabhighlight}4  &\cellcolor{tabhighlight}66.26\% &\cellcolor{tabhighlight}96.20\% &\cellcolor{tabhighlight}30.00\% &\cellcolor{tabhighlight}\textbf{68.63\%} &\cellcolor{tabhighlight}99.08\% &\cellcolor{tabhighlight}94.22\% &\cellcolor{tabhighlight}75.73\%  \\

\cline{2-12} 

& \multirow{3}{*}{Flow} 
& VACE~\cite{jiang2025vace} &LTX-0.9B  &80  &54.38\%  &\textbf{95.19\%}  &70.00\%  &52.16\%  &\textbf{98.21\%}  &91.05\%  &76.83\%  \\
& & VACE~\cite{jiang2025vace} &Wan-14B  &100  &\textbf{59.80\%}  &94.66\%  &\textbf{85.00\%}  &53.41\%  &97.80\%  &\textbf{91.62\%}  &80.38\%  \\
& &\cellcolor{tabhighlight}\methodbf (Ours) & \cellcolor{tabhighlight}Wan-14B  & \cellcolor{tabhighlight}4  &\cellcolor{tabhighlight}59.27\% &\cellcolor{tabhighlight}95.01\% &\cellcolor{tabhighlight}\textbf{85.00\%} &\cellcolor{tabhighlight}\textbf{55.75\%} &\cellcolor{tabhighlight}97.92\% &\cellcolor{tabhighlight}91.58\% &\cellcolor{tabhighlight}\textbf{80.75\%}  \\

\cline{2-12} 

& \multirow{3}{*}{Scribble} 
& VACE~\cite{jiang2025vace} &LTX-0.9B  &80  &55.16\%  &96.57\%  &35.00\%  &66.52\%  &\textbf{99.05\%}  &94.36\%  &74.44\%  \\
& & VACE~\cite{jiang2025vace} &Wan-14B  &100  &\textbf{58.26\%}  &\textbf{96.78\%}  &\textbf{50.00\%}  &66.91\%  &98.91\%  &94.89\%  & 77.63\%  \\
& &\cellcolor{tabhighlight}\methodbf (Ours) & \cellcolor{tabhighlight}Wan-14B  & \cellcolor{tabhighlight}4  &\cellcolor{tabhighlight}57.54\% &\cellcolor{tabhighlight}96.64\% &\cellcolor{tabhighlight}\textbf{50.00\%} &\cellcolor{tabhighlight}\textbf{68.23\%} &\cellcolor{tabhighlight}98.97\% &\cellcolor{tabhighlight}\textbf{94.94\%} &\cellcolor{tabhighlight}\textbf{77.72\%}  \\

\cline{2-12} 

& \multirow{3}{*}{Grey} 
& VACE~\cite{jiang2025vace} &LTX-0.9B  &80  &66.45\%  &97.97\%  &10.00\%  &63.60\%  &\textbf{99.34\%}  &97.85\%  &72.54\%  \\
& & VACE~\cite{jiang2025vace} &Wan-14B  &100  &68.87\%  &\textbf{98.25\%}  &15.00\%  &\textbf{66.11\%}  &99.30\%  &98.02\%  & 74.25\%  \\
& &\cellcolor{tabhighlight}\methodbf (Ours) & \cellcolor{tabhighlight}Wan-14B  & \cellcolor{tabhighlight}4  &\cellcolor{tabhighlight}\textbf{69.08\%} &\cellcolor{tabhighlight}98.01\% &\cellcolor{tabhighlight}\textbf{20.00\%} &\cellcolor{tabhighlight}65.92\% &\cellcolor{tabhighlight}99.30\% &\cellcolor{tabhighlight}\textbf{98.08\%}  &\cellcolor{tabhighlight}\textbf{75.07\% } \\

\midrule 

\multirow{15}{*}{Reference}& \multirow{3}{*}{Depth} 
& VACE~\cite{jiang2025vace} &LTX-0.9B  &80  &52.92\%  &87.63\%  &60.00\%  &63.22\%  &\textbf{98.82\%}  &77.83\%  &73.40\%  \\
& & VACE~\cite{jiang2025vace} &Wan-14B  &100  &\textbf{70.07\%}  &96.53\%  &\textbf{65.00\%}  &67.61\%  &98.39\%  &94.97\%  & \textbf{82.09\%}  \\
& &\cellcolor{tabhighlight}\methodbf (Ours) & \cellcolor{tabhighlight}Wan-14B  & \cellcolor{tabhighlight}4  &\cellcolor{tabhighlight}69.49\% &\cellcolor{tabhighlight}\textbf{96.64\%} &\cellcolor{tabhighlight}\textbf{65.00\%} &\cellcolor{tabhighlight}\textbf{67.85\%} &\cellcolor{tabhighlight}98.44\% &\cellcolor{tabhighlight}\textbf{95.05\%} &\cellcolor{tabhighlight}82.08\%  \\

\cline{2-12} 

& \multirow{3}{*}{Pose} 
& VACE~\cite{jiang2025vace} &LTX-0.9B  &80  &49.89\%  &85.24\%  &65.00\%  &52.74\%  &\textbf{98.43\%}  &71.46\%  &70.46\%  \\
& & VACE~\cite{jiang2025vace} &Wan-14B  &100  &\textbf{71.05\%}  &95.38\%  &\textbf{75.00\%}  &67.72\%  &97.77\%  &93.40\%  & 83.38\%  \\
& &\cellcolor{tabhighlight}\methodbf (Ours) & \cellcolor{tabhighlight}Wan-14B  & \cellcolor{tabhighlight}4  &\cellcolor{tabhighlight}70.81\% &\cellcolor{tabhighlight}\textbf{95.97\%} &\cellcolor{tabhighlight}\textbf{75.00\%} &\cellcolor{tabhighlight}\textbf{67.81\%} &\cellcolor{tabhighlight}98.19\% &\cellcolor{tabhighlight}\textbf{94.24\%} &\cellcolor{tabhighlight}\textbf{83.67\%}  \\

\cline{2-12} 

& \multirow{3}{*}{Flow} 
& VACE~\cite{jiang2025vace} &LTX-0.9B  &80  &52.33\%  &83.37\%  &85.00\%  &58.50\%  &\textbf{98.36\%}  &66.36\%  &73.98\%  \\
& & VACE~\cite{jiang2025vace} &Wan-14B  &100  &\textbf{65.92\%}  &96.30\%  &\textbf{90.00\%}  &\textbf{66.74\%}  &97.71\%  &93.01\%  & \textbf{84.94\%}  \\
& &\cellcolor{tabhighlight}\methodbf (Ours) & \cellcolor{tabhighlight}Wan-14B  & \cellcolor{tabhighlight}4  &\cellcolor{tabhighlight}65.31\% &\cellcolor{tabhighlight}\textbf{96.67\%} &\cellcolor{tabhighlight}80.00\% &\cellcolor{tabhighlight}66.54\% &\cellcolor{tabhighlight}98.00\% &\cellcolor{tabhighlight}\textbf{93.78\%} &\cellcolor{tabhighlight}83.38\%  \\

\cline{2-12} 

& \multirow{3}{*}{Scribble} 
& VACE~\cite{jiang2025vace} &LTX-0.9B  &80  &51.64\%  &89.42\%  &50.00\%  &56.96\%  &\textbf{98.70\%}  &73.25\%  &69.99\%  \\
& & VACE~\cite{jiang2025vace} &Wan-14B  &100  &\textbf{65.57\%}  &96.43\%  &\textbf{75.00\%}  &\textbf{66.20\%}  &98.19\%  &93.18\%  & \textbf{82.42\%}  \\
& &\cellcolor{tabhighlight}\methodbf (Ours) & \cellcolor{tabhighlight}Wan-14B  & \cellcolor{tabhighlight}4  &\cellcolor{tabhighlight}64.10\% &\cellcolor{tabhighlight}\textbf{96.78\%} &\cellcolor{tabhighlight}\textbf{75.00\%} &\cellcolor{tabhighlight}65.35\% &\cellcolor{tabhighlight}98.30\% &\cellcolor{tabhighlight}\textbf{93.41\%} &\cellcolor{tabhighlight}82.16\%  \\

\cline{2-12} 

& \multirow{3}{*}{Grey} 
& VACE~\cite{jiang2025vace} &LTX-0.9B  &80  &58.04\%  &95.46\%  &25.00\%  &60.91\%  &\textbf{99.06\%}  &95.68\%  &72.35\%  \\
& & VACE~\cite{jiang2025vace} &Wan-14B  &100  &\textbf{72.00\%}  &97.60\%  &\textbf{30.00\%}  &64.43\%  &98.75\%  &97.45\%  & 76.70\%  \\
& &\cellcolor{tabhighlight}\methodbf (Ours) & \cellcolor{tabhighlight}Wan-14B  & \cellcolor{tabhighlight}4  &\cellcolor{tabhighlight}71.98\% &\cellcolor{tabhighlight}\textbf{97.66\%} &\cellcolor{tabhighlight}\textbf{30.00\%} &\cellcolor{tabhighlight}\textbf{64.52\%} &\cellcolor{tabhighlight}98.81\% &\cellcolor{tabhighlight}\textbf{97.48\%} &\cellcolor{tabhighlight}\textbf{76.74\%}  \\

\bottomrule
\end{tabular}
\label{tab:comp2}
\vspace{10pt}
\end{table*}

\begin{figure*}[h!]
    \centering
    \subfloat{\includegraphics[width=1.0\linewidth]{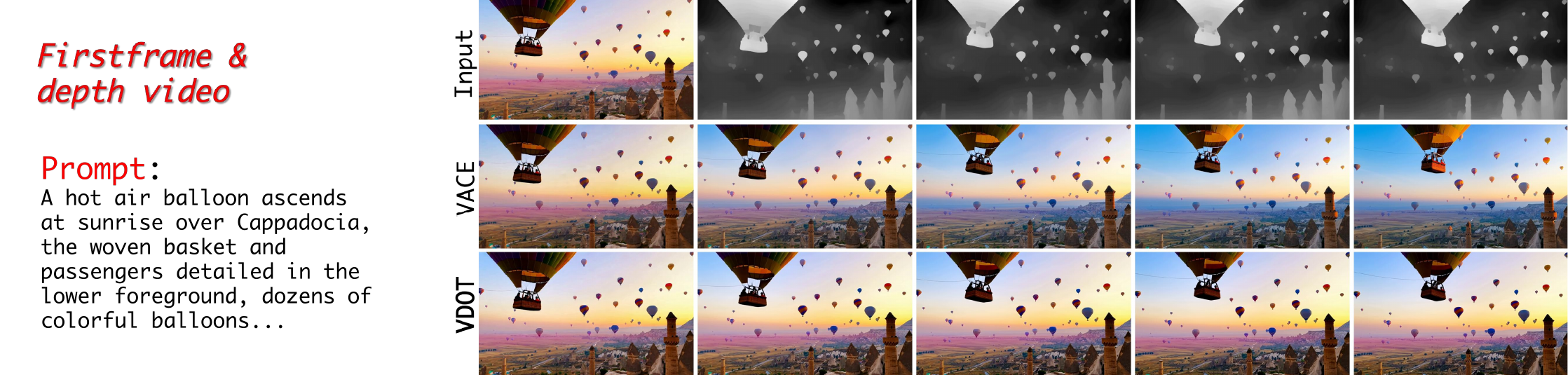}}\\
    \subfloat{\includegraphics[width=1.0\linewidth]{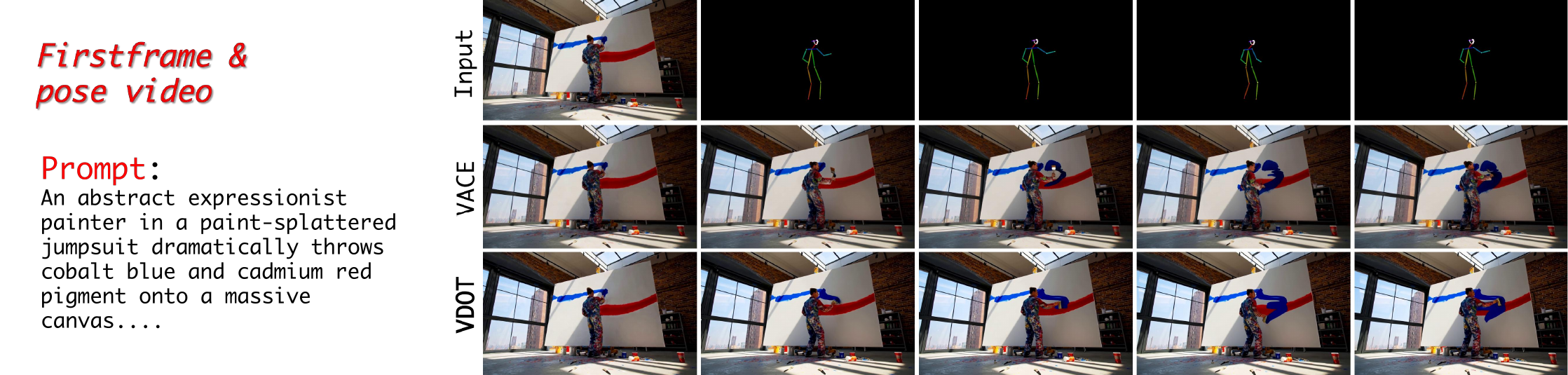}} \\
    \subfloat{\includegraphics[width=1.0\linewidth]{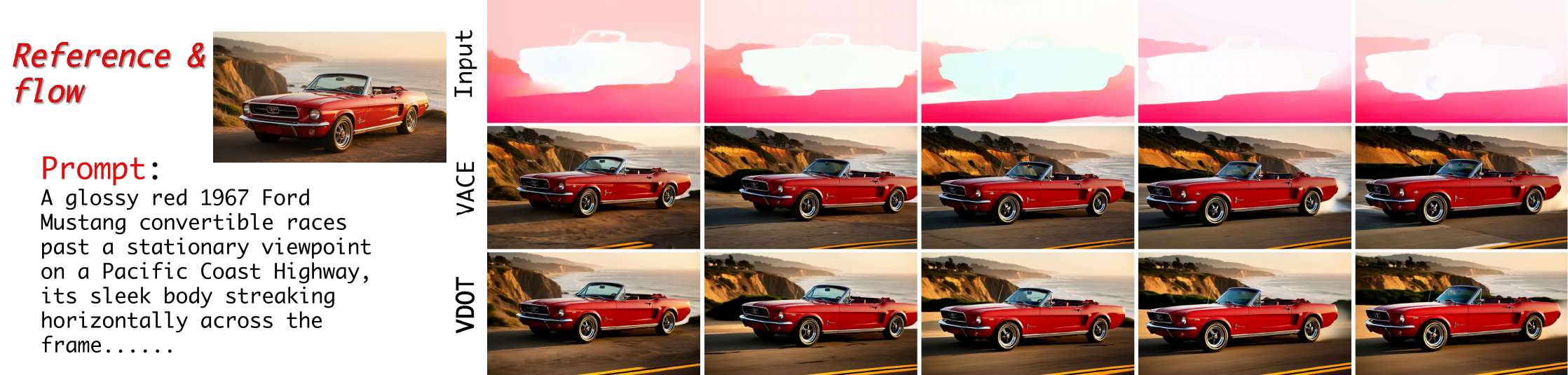}} \\
    \subfloat{\includegraphics[width=1.0\linewidth]{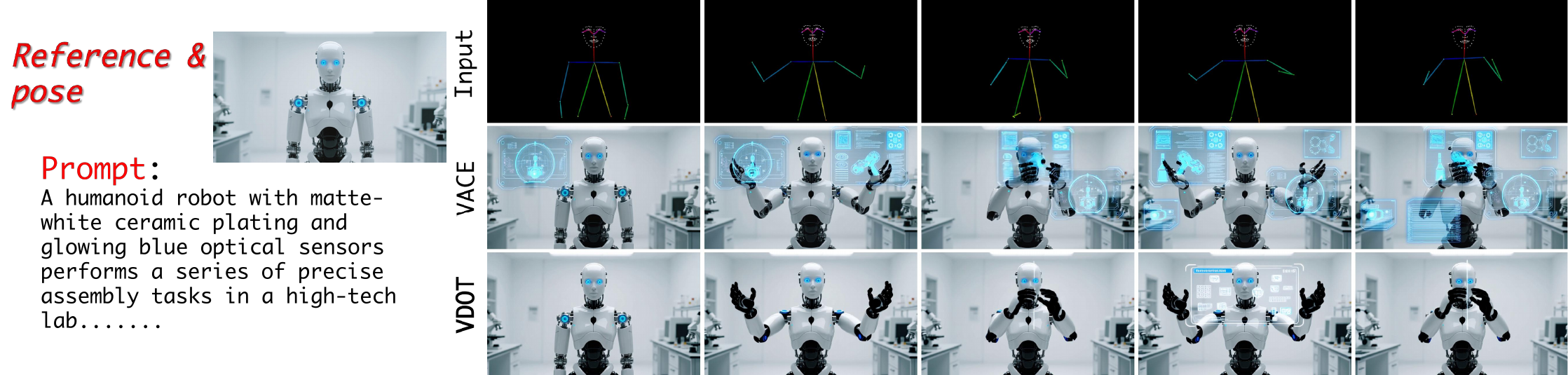}} \\
    \subfloat{\includegraphics[width=1.0\linewidth]{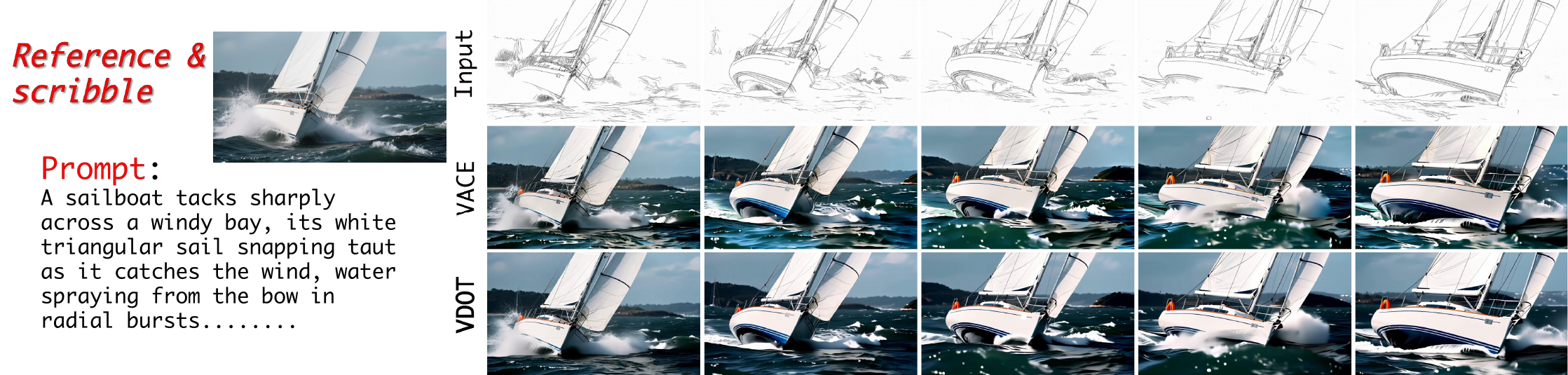}}
    \caption{\textbf{Qualitative comparison of composite tasks.}}
    \label{fig:composite}
\end{figure*}

\noindent\textbf{Composite tasks.} 
Table~\ref{tab:comp2} presents the quantitative evaluation of composite tasks on the \methodbench. 
Across all settings, our proposed \method consistently matches or outperforms VACE in terms of both video quality and temporal consistency, while requiring only 4 NFEs. 
Notably, \method achieves the highest normalized average score in 6 out of 10 settings, showing substantial gains in \emph{Imaging Quality} and \emph{Subject Consistency}. 
In conditions where structure guidance is crucial (e.g., \emph{Depth}, \emph{Scribble}, \emph{Grey}), \method delivers robust improvements over the VACE baselines, demonstrating superior temporal stability and perceptual fidelity. 
Figure~\ref{fig:composite} provides a qualitative comparison between \method and VACE. 
We visualize frames at uniformly spaced intervals (indices 1, 21, 41, 61, and 81). 
In the \emph{firstframe\&depth} setting, VACE suffers from inconsistent background coloration (e.g., in the sky), whereas \method maintains strong spatiotemporal consistency. 
In the \emph{Reference\&pose} setting, also known as the animate anyone setting, both VACE and \method adhere well to the input pose skeleton; however, minor discrepancies appear in the generated effects. 
Specifically, our generated effects align more closely with the motion, adhering to developmental patterns.

\begin{table*}[t]
\caption{
\textbf{Quantitative evaluations on VACE-Benchmark}. We compare the automated score metrics of our \method with VACE on the dimensions of video quality and video consistency.
}

\centering
\footnotesize
\setlength{\tabcolsep}{6.5pt}
\begin{tabular}{l|l|c|c|cccccc|c}

\toprule

Type & Method & Base Model & \#NFE$\downarrow$ & \multicolumn{7}{c}{\textbf{Video Quality \& Video Consistency}} \\

& & & & \rotatebox{90}{\begin{minipage}{1.8cm} {\tikz\fill[video_quality] (0,0) circle (.5ex);} \centering Aesthetic Quality\end{minipage}}
& \rotatebox{90}{\begin{minipage}{1.8cm} {\tikz\fill[video_quality] (0,0) circle (.5ex);} \centering Background Consistency\end{minipage}}
& \rotatebox{90}{\begin{minipage}{1.8cm} {\tikz\fill[video_quality] (0,0) circle (.5ex);} \centering Dynamic Degree\end{minipage}}
& \rotatebox{90}{\begin{minipage}{1.8cm} {\tikz\fill[video_quality] (0,0) circle (.5ex);} \centering Imaging Quality\end{minipage}}
& \rotatebox{90}{\begin{minipage}{1.8cm} {\tikz\fill[video_quality] (0,0) circle (.5ex);} \centering Motion Smoothness\end{minipage}}
& \rotatebox{90}{\begin{minipage}{1.8cm} {\tikz\fill[video_quality] (0,0) circle (.5ex);} \centering Subject Consistency\end{minipage}}
& \rotatebox{90}{\begin{minipage}{1.8cm} {\tikz\fill[score_average] (0,0) circle (.5ex);} \centering Normalized Average\end{minipage}} \\

\midrule 

\multirow{3}{*}{Depth} 
& VACE~\cite{jiang2025vace} &LTX-0.9B  &80  &55.73\%  &96.06\%  &60.00\%  &67.53\%  &\textbf{98.92\%}  &93.90\%  &78.69\%  \\
& VACE~\cite{jiang2025vace} &Wan-14B  &100  &62.34\%  &\textbf{96.10\%}  &\textbf{65.00\%}  &68.85\%  &98.57\%  &94.25\%  &80.85\%  \\
&\cellcolor{tabhighlight}\methodbf (Ours) & \cellcolor{tabhighlight}Wan-14B  &\cellcolor{tabhighlight}4 &\cellcolor{tabhighlight}\textbf{62.46\%} & \cellcolor{tabhighlight}95.81\% & \cellcolor{tabhighlight}\textbf{65.00\%} & \cellcolor{tabhighlight}\textbf{69.69\%} & \cellcolor{tabhighlight}98.39\% & \cellcolor{tabhighlight}\textbf{94.53\%} &\cellcolor{tabhighlight}\textbf{80.98\%}  \\

\midrule 

\multirow{3}{*}{Pose} 
& VACE~\cite{jiang2025vace} &LTX-0.9B  &80  &59.97\%  &94.80\%  &\textbf{85.00\%}  &66.85\%  &\textbf{98.93\%}  &\textbf{94.91\%}  &83.41\%  \\
& VACE~\cite{jiang2025vace} &Wan-14B  &100  &66.69\%  &\textbf{95.32\%}  &75.00\%  &68.48\%  &98.56\%  &94.19\%  &83.04\%  \\
&\cellcolor{tabhighlight}\methodbf (Ours) & \cellcolor{tabhighlight}Wan-14B  &\cellcolor{tabhighlight}4  
&\cellcolor{tabhighlight}\textbf{66.71\%} & \cellcolor{tabhighlight}94.99\% & \cellcolor{tabhighlight}80.00\% & \cellcolor{tabhighlight}\textbf{70.40\%} &\cellcolor{tabhighlight}98.45\% &\cellcolor{tabhighlight}94.26\% &\cellcolor{tabhighlight}\textbf{84.13\%}  \\

\midrule 

\multirow{3}{*}{Flow} 
& VACE~\cite{jiang2025vace} &LTX-0.9B  &80  &55.36\%  &95.99\%  &\textbf{75.00\%}  &64.30\%  &\textbf{99.07\%}  &94.14\%  &80.64\%  \\
& VACE~\cite{jiang2025vace} &Wan-14B  &100  &\textbf{60.44\%}  &\textbf{96.09\%}  &\textbf{75.00\%}  &66.99\%  &98.48\%  &\textbf{94.71\%}  &81.95\%  \\
&\cellcolor{tabhighlight}\methodbf (Ours) & \cellcolor{tabhighlight}Wan-14B  &\cellcolor{tabhighlight}4  &\cellcolor{tabhighlight}60.24\% & \cellcolor{tabhighlight}95.65\% & \cellcolor{tabhighlight}\textbf{75.00\%} & \cellcolor{tabhighlight}\textbf{70.02\%} & \cellcolor{tabhighlight}98.42\% & \cellcolor{tabhighlight}94.26\% &\cellcolor{tabhighlight}\textbf{82.27\%}  \\
 
\midrule 

\multirow{3}{*}{Scribble} 
& VACE~\cite{jiang2025vace} &LTX-0.9B  &80  &53.81\%  &96.57\%  &45.00\%  &66.05\%  &\textbf{99.21\%}  &95.23\%  &75.98\%  \\
& VACE~\cite{jiang2025vace} &Wan-14B  &100  &\textbf{59.44\%}  &96.49\%  &\textbf{50.00\%}  &67.01\%  &98.46\%  &95.44\%  &77.80\%  \\
&\cellcolor{tabhighlight}\methodbf (Ours) & \cellcolor{tabhighlight}Wan-14B  &\cellcolor{tabhighlight}4  &\cellcolor{tabhighlight}57.99\% & \cellcolor{tabhighlight}\textbf{96.75\%} & \cellcolor{tabhighlight}\textbf{50.00\%} & \cellcolor{tabhighlight}\textbf{69.16\%} & \cellcolor{tabhighlight}98.60\% & \cellcolor{tabhighlight}\textbf{95.79\%} &\cellcolor{tabhighlight}\textbf{78.05\%} \\

\midrule 

\multirow{3}{*}{Grey} 
& VACE~\cite{jiang2025vace} &LTX-0.9B  &80  &58.63\%  &\textbf{95.65\%}  &55.00\%  &59.93\%  &\textbf{98.89\%}  &92.39\%  &76.75\%  \\
& VACE~\cite{jiang2025vace} &Wan-14B  &100 &\textbf{62.87\%}  &95.46\%  &60.00\%  &66.35\%  &98.40\%  &92.91\%  &79.33\%  \\
&\cellcolor{tabhighlight}\methodbf (Ours) & \cellcolor{tabhighlight}Wan-14B  &\cellcolor{tabhighlight}4  &\cellcolor{tabhighlight}62.25\% & \cellcolor{tabhighlight}95.64\% & \cellcolor{tabhighlight}\textbf{65.00\%} & \cellcolor{tabhighlight}\textbf{66.74\%} & \cellcolor{tabhighlight}98.40\% & \cellcolor{tabhighlight}\textbf{92.88\%} &\cellcolor{tabhighlight}\textbf{80.15\%}  \\

\midrule 

\multirow{3}{*}{Extension} 
& VACE~\cite{jiang2025vace} &LTX-0.9B  &80  &57.82\%  &\textbf{96.06\%}  &35.00\%  &68.15\%  &\textbf{99.34\%}  &\textbf{94.06\%}  &75.07\%  \\
& VACE~\cite{jiang2025vace} &Wan-14B  &100  &\textbf{61.34\%}  &95.05\%  &\textbf{60.00\%}  &69.55\%  &98.75\%  &92.83\%  &\textbf{79.59\%}  \\
&\cellcolor{tabhighlight}\methodbf (Ours) & \cellcolor{tabhighlight}Wan-14B  &\cellcolor{tabhighlight}4  &\cellcolor{tabhighlight}61.25\% & \cellcolor{tabhighlight}95.94\% & \cellcolor{tabhighlight}45.00\% & \cellcolor{tabhighlight}\textbf{70.01\%} & \cellcolor{tabhighlight}99.04\% & \cellcolor{tabhighlight}93.86\% &\cellcolor{tabhighlight}77.52\%  \\

\midrule 

\multirow{3}{*}{Outpaint} 
& VACE~\cite{jiang2025vace} &LTX-0.9B  &80  &56.98\%  &96.47\%  &45.00\%  &\textbf{69.39\%}  &\textbf{99.17\%}  &\textbf{95.50\%}  &77.08\%  \\
& VACE~\cite{jiang2025vace} &Wan-14B  &100 &58.53\%  &96.83\%  &40.00\%  &68.55\%  &99.03\%  &95.34\%  &76.38\%  \\
&\cellcolor{tabhighlight}\methodbf (Ours) & \cellcolor{tabhighlight}Wan-14B  &\cellcolor{tabhighlight}4  &\cellcolor{tabhighlight}\textbf{58.86\%} & \cellcolor{tabhighlight}\textbf{96.84\%} & \cellcolor{tabhighlight}\textbf{50.00\%} & \cellcolor{tabhighlight}69.18\% & \cellcolor{tabhighlight}99.00\% & \cellcolor{tabhighlight}95.48\% &\cellcolor{tabhighlight}\textbf{78.23\%}  \\

\midrule 

\multirow{3}{*}{R2V} 
& VACE~\cite{jiang2025vace} &LTX-0.9B  &80 &62.57\%  &97.85\%  &17.50\%  &\textbf{72.32\%}  &\textbf{99.47\%}  &97.91\%  &74.60\%  \\
& VACE~\cite{jiang2025vace} &Wan-14B  &100 &\textbf{65.99\%}  &\textbf{98.77\%}  &15.00\%  &71.19\%  &99.38\%  &\textbf{98.67\%}  &74.83\%  \\
&\cellcolor{tabhighlight}\methodbf (Ours) & \cellcolor{tabhighlight}Wan-14B  &\cellcolor{tabhighlight}4  &\cellcolor{tabhighlight}65.47\% & \cellcolor{tabhighlight}98.58\% & \cellcolor{tabhighlight}\textbf{20.00\%} & \cellcolor{tabhighlight}71.94\% & \cellcolor{tabhighlight}99.32\% & \cellcolor{tabhighlight}98.48\% &\cellcolor{tabhighlight}\textbf{75.63\%}  \\

\midrule 

\multirow{3}{*}{T2V} 
& VACE~\cite{jiang2025vace} &LTX-0.9B  &80  &58.08\%  &\textbf{97.94\%}  &40.00\%  &\textbf{71.16\%}  &\textbf{99.06\%}  &\textbf{97.71\%}  &77.33\%  \\
& VACE~\cite{jiang2025vace} &Wan-14B  &100  &63.49\%  &97.63\%  &\textbf{45.00\%}  &69.49\%  &98.75\%  &96.96\%  &\textbf{78.55\%}  \\
&\cellcolor{tabhighlight}\methodbf (Ours) & \cellcolor{tabhighlight}Wan-14B  &\cellcolor{tabhighlight}4  &\cellcolor{tabhighlight}\textbf{64.37\%} & \cellcolor{tabhighlight}97.75\% & \cellcolor{tabhighlight}35.00\% & \cellcolor{tabhighlight}70.65\% & \cellcolor{tabhighlight}99.02\% & \cellcolor{tabhighlight}97.30\% &\cellcolor{tabhighlight}77.35\%  \\

\bottomrule
\end{tabular}
\label{tab:comp_vacebench}
\end{table*}
\noindent\textbf{Results of VACE-Benchmark.} 
Quantitative evaluations on the VACE-Benchmark demonstrate that our proposed \method achieves a superior balance between generation quality and computational efficiency compared to the VACE baselines. 
As shown in Table~\ref{tab:comp_vacebench}, our method consistently attains comparable or higher scores across diverse tasks, such as \emph{pose}, \emph{depth}, and \emph{flow}, particularly in dimensions of \emph{Imaging Quality} and \emph{Subject Consistency}. 
The substantial reduction in inference steps, combined with top-tier performance across the majority of tasks, highlights our method's ability to maintain high video fidelity and temporal stability with minimal inference latency.

\begin{figure*}[h!]
    \centering
    \subfloat{\includegraphics[width=1.0\linewidth]{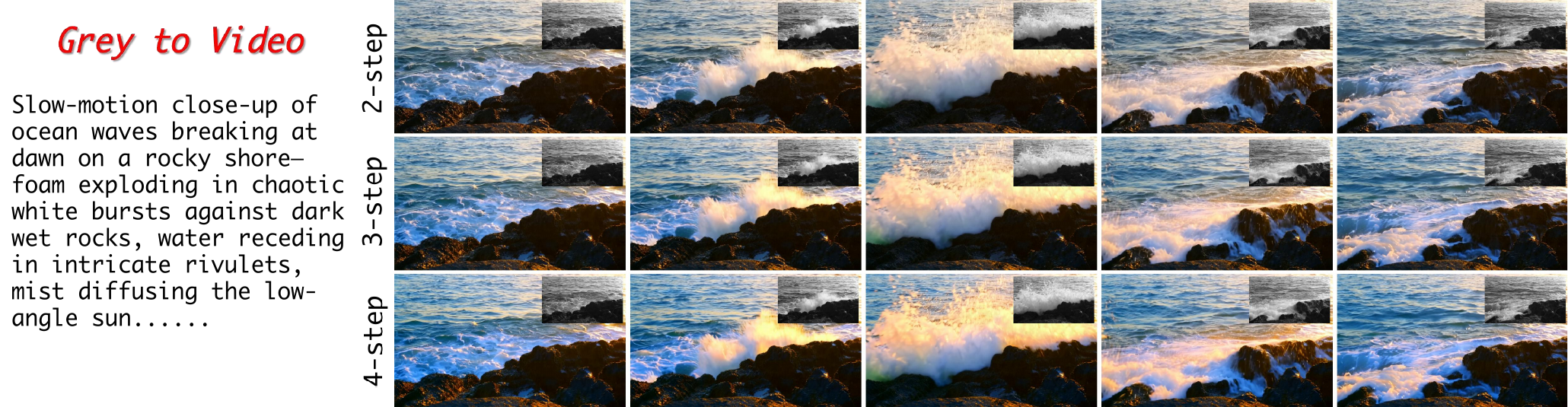}}\\
    \subfloat{\includegraphics[width=1.0\linewidth]{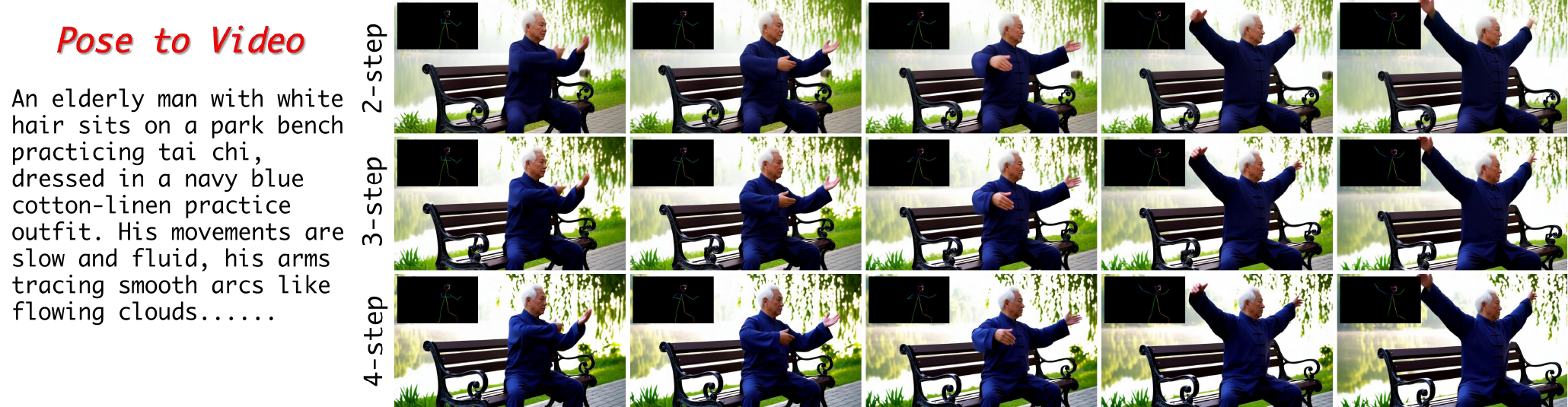}} 
    \caption{\textbf{Qualitative comparison of different denoising steps of \method.}}
    \label{fig:steps}
\end{figure*}

\begin{table*}[t]
\caption{
\textbf{Quantitative comparison of denosing steps of \method on \methodbench}.
}
\centering
\setlength{\tabcolsep}{5.5pt}
\begin{tabular}{c|cccccccc}
\toprule
\textbf{Denoising Steps} & \textbf{Depth} & \textbf{Pose} & \textbf{Flow} & \textbf{Scribble} & \textbf{Grey} & \textbf{Outpaint} & \textbf{Extension} & \textbf{R2V} \\ \midrule 
\textbf{2 steps} &77.82\%  &79.65\%  &78.08\%  &76.82\%  &71.99\%  &75.86\%  &74.83\%  &79.18\%    \\ \hline
\textbf{3 steps} &78.42\%  &80.21\%  &79.09\%  &76.95\%  &72.05\%  &75.70\%  &78.51\%  &80.80\%    \\ \hline
\textbf{4 steps} &78.50\%  &80.54\%  &80.18\%  &76.77\%  &72.13\%  &76.54\%  &80.53\%  &81.32\%    \\ 
\bottomrule
\end{tabular}
\label{tab:steps}
\end{table*}
\noindent\textbf{Denoising steps.} 
Our training adheres to a four-step paradigm based on Self-Forcing~\cite{huang2025self}. 
Evaluations on \methodbench reveal the impact of step counts on generation quality. 
Table~\ref{tab:steps} shows that increasing denoising steps consistently boosts performance, particularly for \emph{pose} and \emph{extension} tasks. 
This can be attributed to the coarse-to-fine nature of the generation process: low-frequency structure is determined early, while fine-grained details require more steps to resolve. 
Thus, tasks providing strong structural constraints (e.g., \emph{scribble}, \emph{grey}) saturate quickly, with the 2-step model performing similarly to the 4-step model. 
In contrast, under-constrained tasks like \emph{pose} and \emph{extension} benefit substantially from additional steps. 
Visualizations in Figure~\ref{fig:steps} validate this observation; for instance, in the pose-to-video task, the 4-step model generates superior fine-grained details (e.g., hair, hands, and background elements) compared to the 2-step output.

\begin{table*}[t]
\caption{
\textbf{Quantitative comparison of generation resolution of \method on \methodbench}.
}
\centering
\setlength{\tabcolsep}{5.5pt}
\begin{tabular}{c|cccccccc}
\toprule
\textbf{Resolution} & \textbf{Depth} & \textbf{Pose} & \textbf{Flow} & \textbf{Scribble} & \textbf{Grey} & \textbf{Outpaint} & \textbf{Extension} & \textbf{R2V} \\ \midrule 
\textbf{832$\times$480} &78.50\%  &80.54\%  &80.18\%  &76.77\%  &72.13\%  &76.54\%  &80.53\%  &81.32\%    \\ \hline
\textbf{1280$\times$720} &78.15\%  &80.34\%  &79.44\%  &76.99\%  &71.88\%  &75.81\%  &77.98\%  &79.15\%    \\ 
\bottomrule
\end{tabular}
\label{tab:resolution}
\end{table*}
\noindent\textbf{Generation resolution.} 
While our model is capable of inference at any resolution, it was trained specifically on $832\times480$ videos. 
Table~\ref{tab:resolution} quantifies the impact of increasing resolution to $1280\times720$, showing a marked decline in performance for most tasks. 
This degradation arises from a domain gap: the model lacks priors for high-resolution details. 
Forced to extrapolate low-resolution features to a larger pixel space, the model generates outputs that suffer from blurred details and noise artifacts.

\begin{figure*}[h!]
    \centering
    \subfloat{\includegraphics[width=1.0\linewidth]{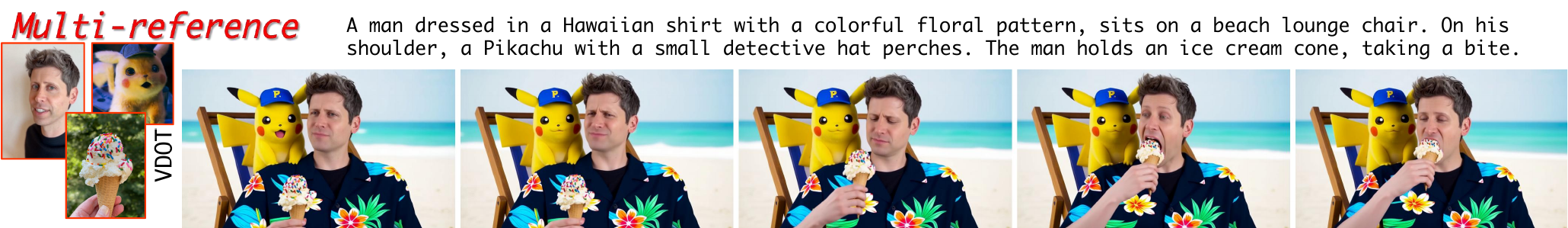}}\\
    \subfloat{\includegraphics[width=1.0\linewidth]{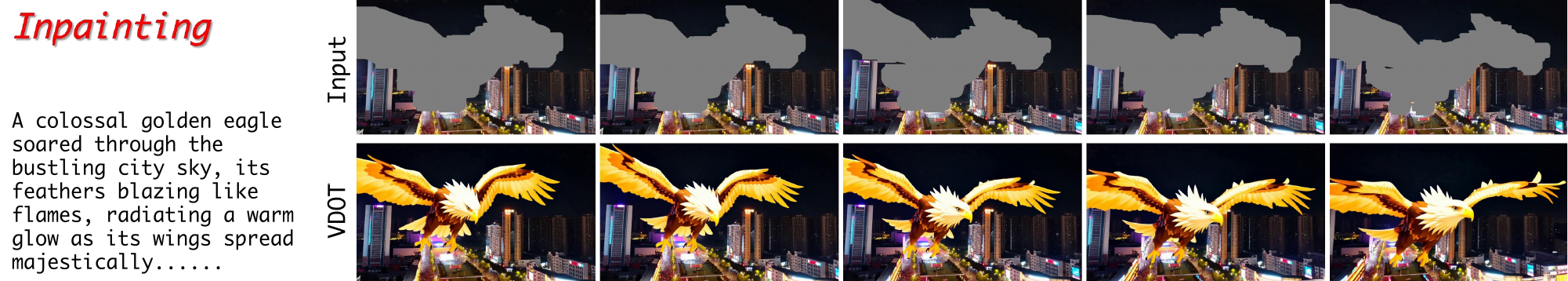}} \\
    \subfloat{\includegraphics[width=1.0\linewidth]{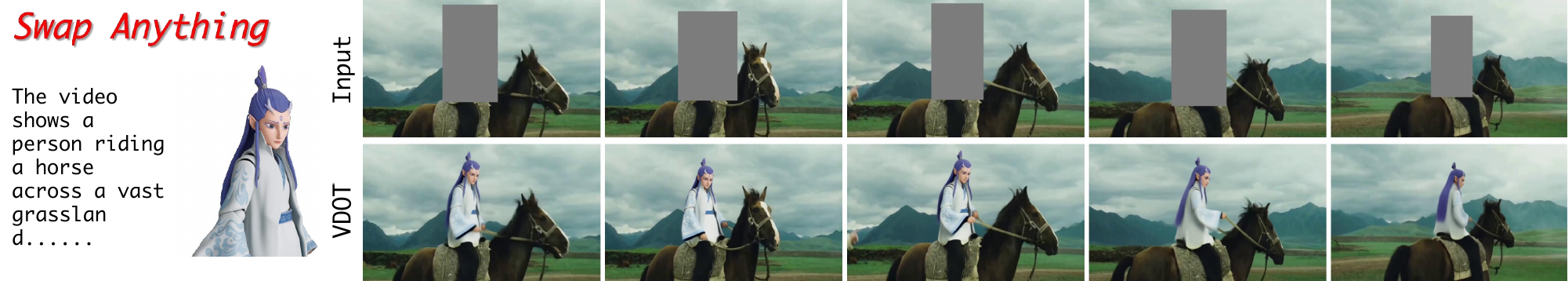}} \\
    \subfloat{\includegraphics[width=1.0\linewidth]{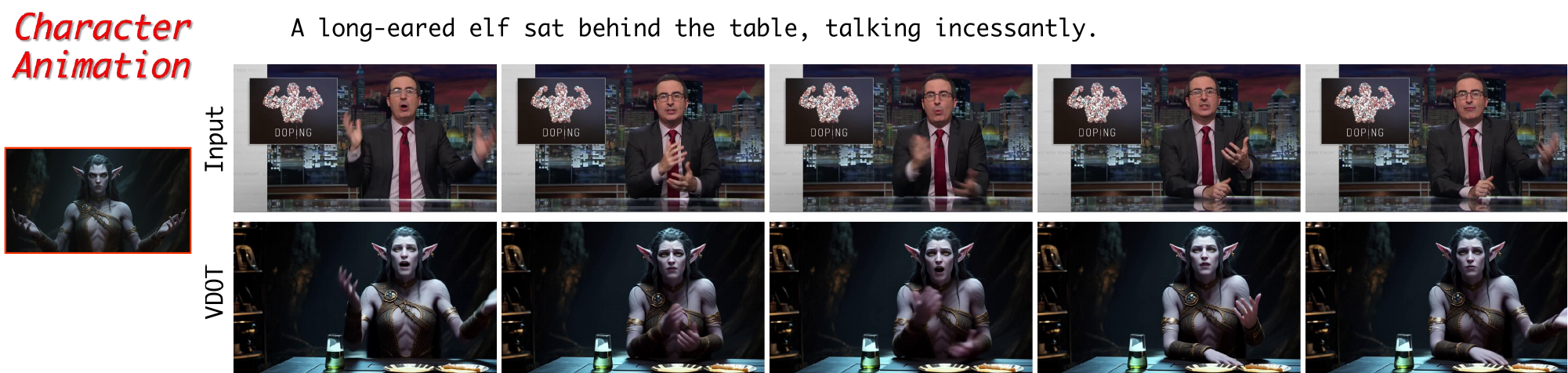}} \\
    \subfloat{\includegraphics[width=1.0\linewidth]{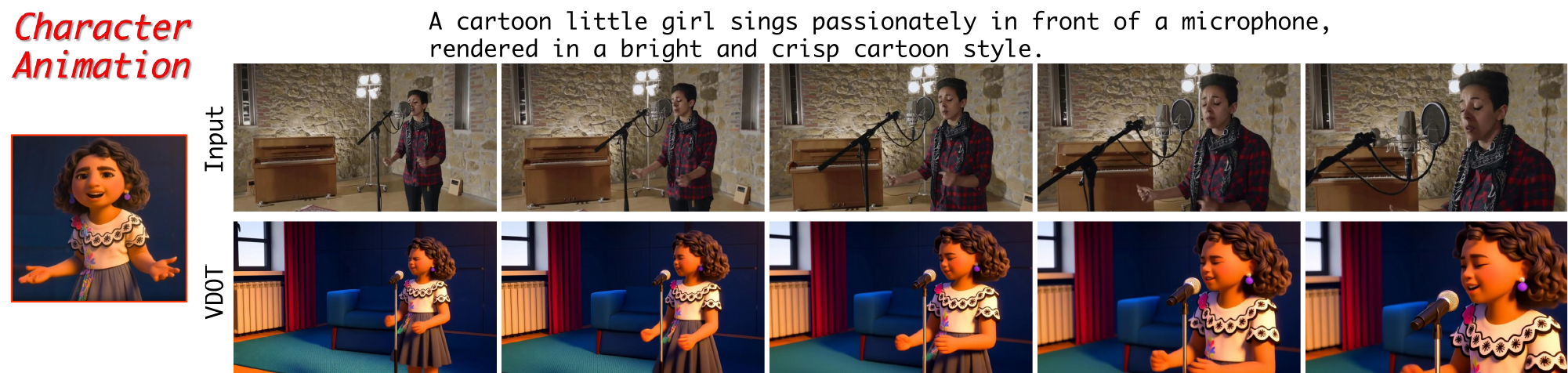}} \\
    \subfloat{\includegraphics[width=1.0\linewidth]{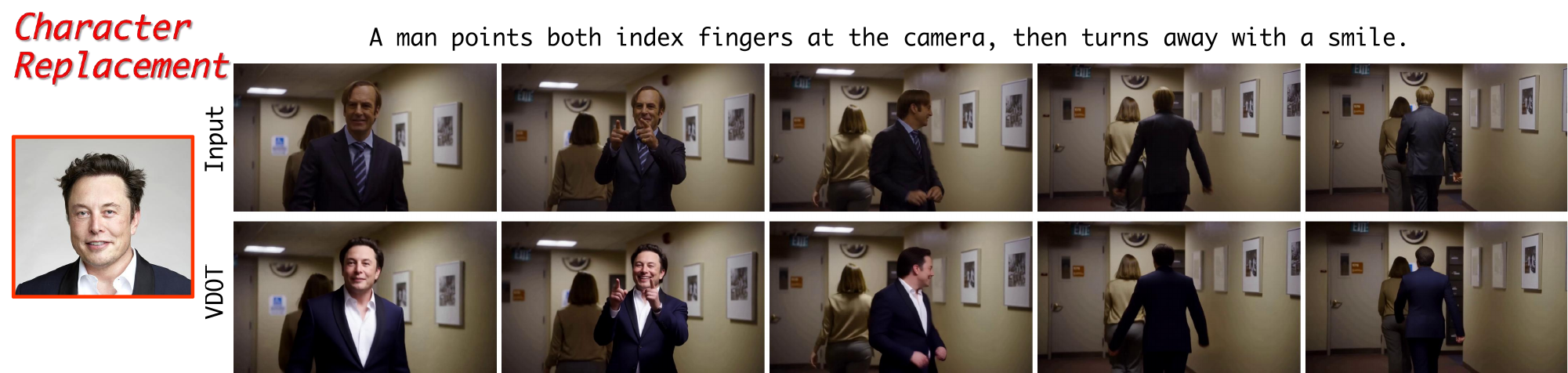}} \\
    \caption{\textbf{Visualization of \method on untrained tasks.}}
    \label{fig:unseen1}
\end{figure*}

\begin{figure*}[h!]
    \centering
    \subfloat{\includegraphics[width=1.0\linewidth]{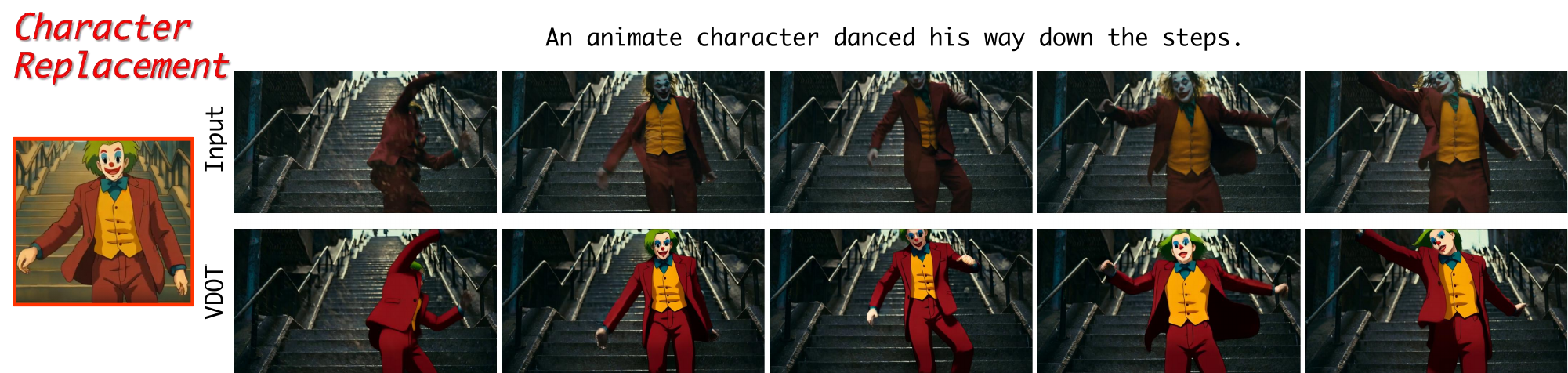}} \\
    \subfloat{\includegraphics[width=1.0\linewidth]{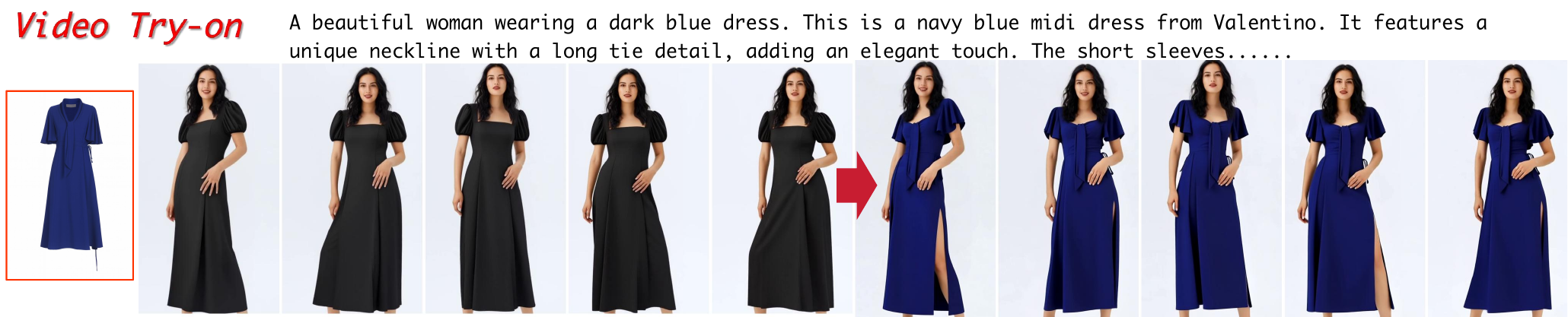}} \\ 
    \subfloat{\includegraphics[width=1.0\linewidth]{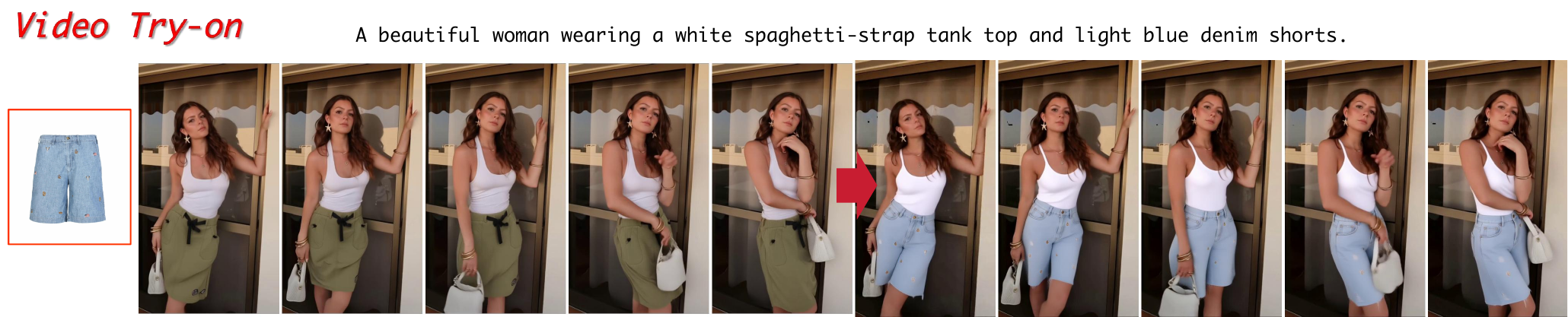}} \\ 
    \subfloat{\includegraphics[width=1.0\linewidth]{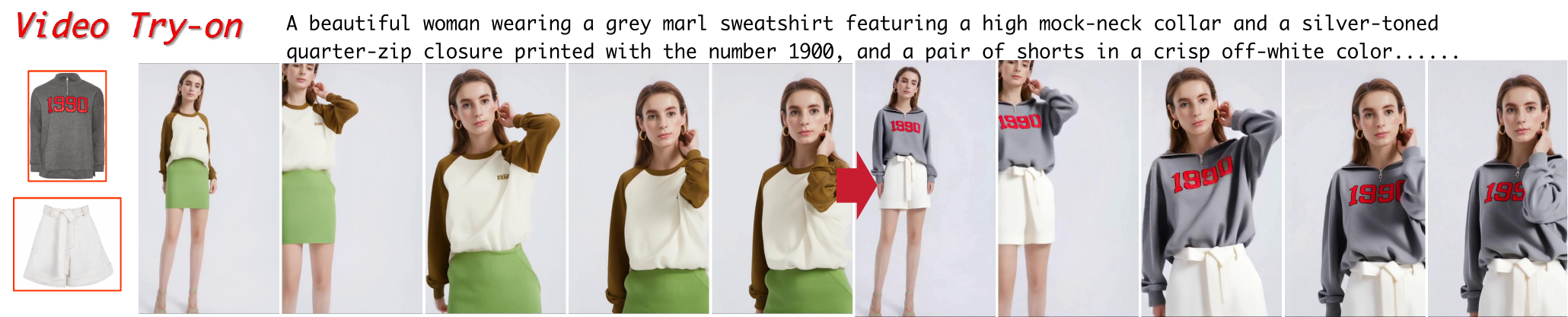}} 
    \caption{\textbf{Visualization of \method on untrained tasks.}}
    \label{fig:unseen2}
\end{figure*}

\noindent\textbf{Generalizability to untrained tasks.} 
Beyond the 18 supervised tasks listed in Table~\ref{tab:training}, our model demonstrates strong potential for few-step generalization. 
Even without specific training, the model can generate high-quality videos for novel tasks using just 4 denoising steps. 
Qualitative results in Figure~\ref{fig:unseen1} and Figure~\ref{fig:unseen2} highlight this versatility, showcasing successful applications in video inpainting, swap anything, character animation, character replacement, and video try-on. 

\end{document}